\documentclass[5p,times]{elsarticle}
\usepackage[switch]{lineno}
\usepackage{hyperref}
\usepackage{comment}
\usepackage{float}
\usepackage{siunitx}

\usepackage[table]{xcolor}

\usepackage{amsmath,amssymb,amsfonts}
\usepackage{kotex}

\usepackage{mathtools}

\usepackage{amsmath}
\usepackage{multirow}
\usepackage{booktabs}
\usepackage{pifont}
\usepackage{xcolor}
\newcommand{\cmark}{\textcolor{green!80!black}{\ding{51}}}
\newcommand{\xmark}{\textcolor{red}{\ding{55}}}

\usepackage{algorithm}
\usepackage{algpseudocode}

\usepackage{graphicx}
\usepackage{textcomp}
\usepackage[caption=false,labelfont=sf,textfont=sf]{subfig} 

\definecolor{mypink}{rgb}{0.0, 0.0, 0.0}
\definecolor{myblue}{rgb}{0.0, 0.0, 0.0} 
\definecolor{mygreen}{rgb}{0.0, 0.0, 0.0} 
\definecolor{mypurple}{rgb}{0.0, 0.0, 0.0} 
\definecolor{myred}{rgb}{0.0, 0.0, 0.0} 
\definecolor{mytealblue}{rgb}{0, 0.0, 0}

\newcommand{\news}[1]{\textcolor{mypink}{#1}} 
\newcommand{\blue}[1]{\textcolor{myblue}{#1}} 
\newcommand{\purple}[1]{\textcolor{mypurple}{#1}} 
\newcommand{\red}[1]{\textcolor{myred}{#1}} 
\newcommand{\tealblue}[1]{\textcolor{mytealblue}{#1}} 

\makeatletter
\def\therule{\makebox[\algorithmicindent][l]{\hspace*{.5em}\vrule height .75\baselineskip depth .25\baselineskip}}%

\newtoks\therules
\therules={}
\def\appendto#1#2{\expandafter#1\expandafter{\the#1#2}}
\def\gobblefirst#1{
  #1\expandafter\expandafter\expandafter{\expandafter\@gobble\the#1}}%
\def\LState{\State\unskip\the\therules}
\def\pushindent{\appendto\therules\therule}%
\def\popindent{\gobblefirst\therules}%
\def\printindent{\unskip\the\therules}%
\def\printandpush{\printindent\pushindent}%
\def\popandprint{\popindent\printindent}%

\algdef{SE}[WHILE]{While}{EndWhile}[1]
  {\printandpush\algorithmicwhile\ #1\ \algorithmicdo}
  {\popandprint\algorithmicend\ \algorithmicwhile}%
\algdef{SE}[FOR]{For}{EndFor}[1]
  {\printandpush\algorithmicfor\ #1\ \algorithmicdo}
  {\popandprint\algorithmicend\ \algorithmicfor}%
\algdef{S}[FOR]{ForAll}[1]
  {\printindent\algorithmicforall\ #1\ \algorithmicdo}%
\algdef{SE}[LOOP]{Loop}{EndLoop}
  {\printandpush\algorithmicloop}
  {\popandprint\algorithmicend\ \algorithmicloop}%
\algdef{SE}[REPEAT]{Repeat}{Until}
  {\printandpush\algorithmicrepeat}[1]
  {\popandprint\algorithmicuntil\ #1}%
\algdef{SE}[IF]{If}{EndIf}[1]
  {\printandpush\algorithmicif\ #1\ \algorithmicthen}
  {\popandprint\algorithmicend\ \algorithmicif}%
\algdef{C}[IF]{IF}{ElsIf}[1]
  {\popandprint\pushindent\algorithmicelse\ \algorithmicif\ #1\ \algorithmicthen}%
\algdef{Ce}[ELSE]{IF}{Else}{EndIf}
  {\popandprint\pushindent\algorithmicelse}%
\algdef{SE}[PROCEDURE]{Procedure}{EndProcedure}[2]
   {\printandpush\algorithmicprocedure\ \textproc{#1}\ifthenelse{\equal{#2}{}}{}{(#2)}}%
   {\popandprint\algorithmicend\ \algorithmicprocedure}%
\algdef{SE}[FUNCTION]{Function}{EndFunction}[2]
   {\printandpush\algorithmicfunction\ \textproc{#1}\ifthenelse{\equal{#2}{}}{}{(#2)}}%
   {\popandprint\algorithmicend\ \algorithmicfunction}%
\makeatother

\usepackage{tikz}
\newcommand{\blackcircle}[1]{%
  \tikz[baseline=(char.base)]{
    \node[shape=circle,draw,inner sep=0.5pt,fill=black,text=white] (char) {#1};
  }
}

\modulolinenumbers[1]

\journal{Future Generation Computer Systems}

\biboptions{sort&compress}

\newcommand{\systemName}{\textsf{QuantuneV2}}

\begin{document}

\begin{frontmatter}

\title{QuantuneV2: Compiler-Based Local Metric-Driven Mixed Precision Quantization for Practical Embedded AI Applications}

\author[mymainaddress]{Jeongseok Kim\corref{contrib}}
\ead{jskim88@kaist.ac.kr}

\author[mysecondaryaddress]{Jemin Lee\corref{contrib}}
\cortext[contrib]{These authors contributed equally to this work and are listed alphabetically.}
\ead{leejaymin@etri.re.kr}

\author[mysecondaryaddress]{Yongin Kwon}
\ead{yongin.kwon@etri.re.kr}

\author[mymainaddress]{Daeyoung Kim\corref{mycorrespondingauthor}}
\cortext[mycorrespondingauthor]{Corresponding author}
\ead{kimd@kaist.ac.kr}

\address[mymainaddress]{School of Computing, Korea Advanced Institute of Science \& Technology (KAIST), Daejeon 34141, South Korea}
\address[mysecondaryaddress]{Artificial Intelligence Computing Research Laboratory, Electronics and Telecommunications Research Institute (ETRI), Daejeon 34129, South Korea}

\begin{abstract}

\news{Mixed-precision quantization methods have been proposed to reduce model size while minimizing accuracy degradation. However, existing studies require retraining and do not consider the computational overhead and intermediate representations (IR) generated during the compilation process, limiting their application at the compiler level. This computational overhead refers to the runtime latency caused by frequent quantization and de-quantization operations during inference. Performing these operations at the individual operator level causes significant runtime delays. To address these issues, we propose {\systemName}, a compiler-based mixed-precision quantization method designed for practical embedded AI applications. \red{{\systemName} performs inference only twice—once before quantization and once after quantization—and operates with a computational complexity off $\mathcal{O}(n)$ that increases linearly with the number of model parameters.} We also made the sensitivity analysis more stable by using local metrics like weights, activation values, the Signal-to-Quantization-Noise Ratio (SQNR), and the Mean Squared Error (MSE). We also cut down on computational overhead by choosing the best IR and using operator fusion. Experimental results show that {\systemName} achieved up to a 10.28\% improvement in accuracy and a 12.52\% increase in speed compared to existing methods across five models: ResNet18v1, ResNet50v1, SqueezeNetv1, VGGNet, and MobileNetv2. This demonstrates that {\systemName} enhances model performance while maintaining computational efficiency, making it suitable for deployment in embedded AI environments.}

\end{abstract}



\end{frontmatter}


\section{Introduction}
\label{sec:introduction}
\begin{figure}
    \centering
    \includegraphics[width=1\linewidth, scale=0.8]{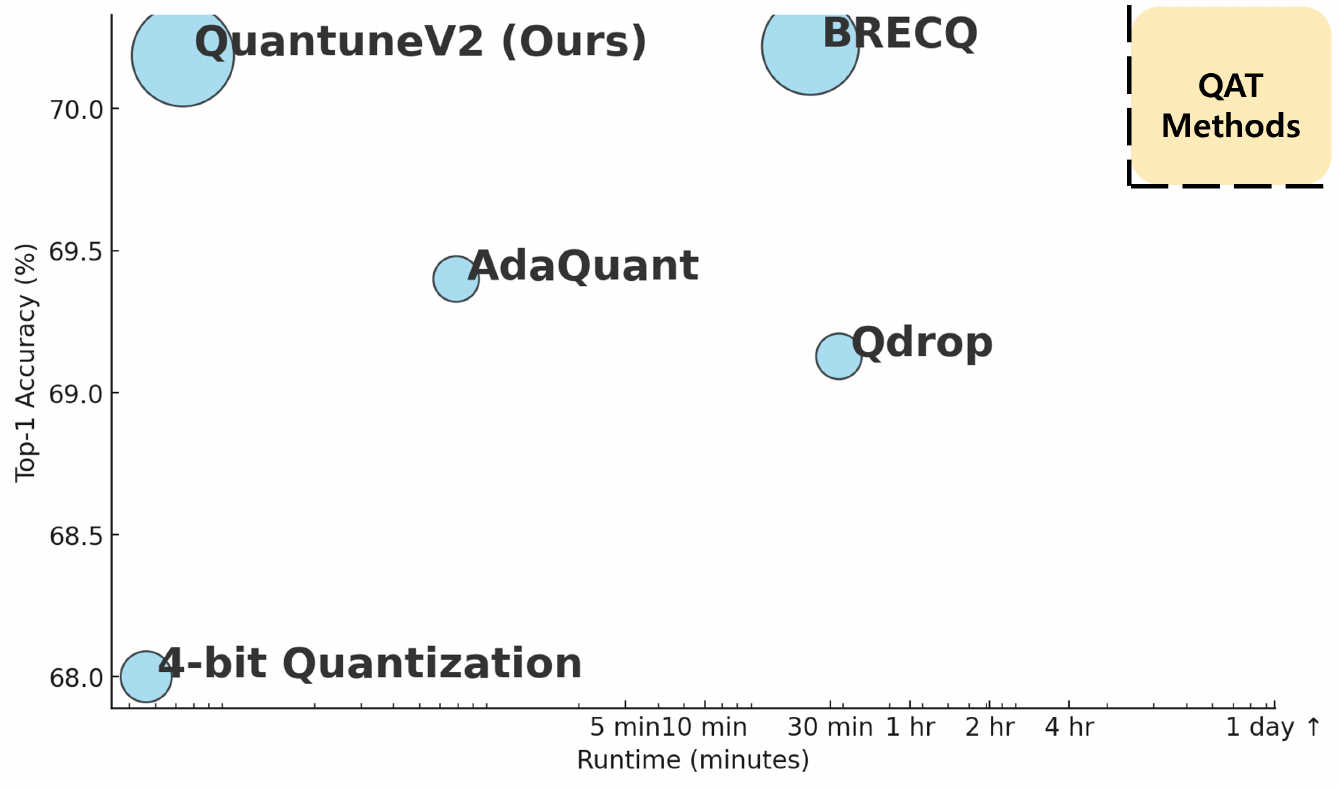}
    \caption{\red{The runtime of different quantization methods for ResNet18v1. We examine PTQ approaches and illustrate the possibilities of orange square (QAT) methods. The size of the circles represents the model size based on the quantization bit-widths.}}
    \label{fig:runtime}
\end{figure}

According to the scaling law~\cite{kaplan2020scaling}, the performance of deep learning models continuously improves as their size increases, leading to a continuous growth in model sizes. With the increase in model size, there has been active research on lightweight methods to deploy deep learning on resource-constrained devices, with quantization being a representative approaches~\cite{chen2020deep}. 
 \tealblue{Compilation time optimization has emerged as a critical challenge in deep learning model development and on-device execution environments, prompting extensive research efforts. In particular, because the time for determining mixed-precision quantization is highly constrained, prolonged compilation can delay the development cycle and decrease productivity~\cite{comopt,mlgo}. Although compilation typically happens only once, frequent model updates or reuse across various environments can make lengthy compilation times severely inefficient. In this context, short compilation for quantization is crucial. As shown in Fig.~\ref{fig:runtime}, {\systemName} demonstrates its effectiveness with minimal runtime overhead.}
Quantization involves converting the weights and activation values of a model from a 32-bit floating-point format to lower bit-widths. Models converted to lower bit-widths operate with reduced memory usage and power consumption. However, uniformly quantizing all layers to lower bit-widths can significantly degrade model accuracy due to loss of precision. For instance, converting layers from 32-bit to 8-bit can result in substantial information loss if the data spans a wide range or includes extreme values.
This information loss is a major cause of accuracy degradation in DNN models~\cite{nagel2021white,Gong_2021}.

To address this issue, many studies have focused on accurately measuring sensitivity and applying mixed-precision quantization based on these measurements to reduce the model size while minimizing accuracy drop~\cite{dong2019hawq,dong2020hawq,yao2021hawq,jacob2017,ewha,pandey2023practical,wu2020integer}.

\begin{figure}
    \centering
    \includegraphics[width=1\linewidth]{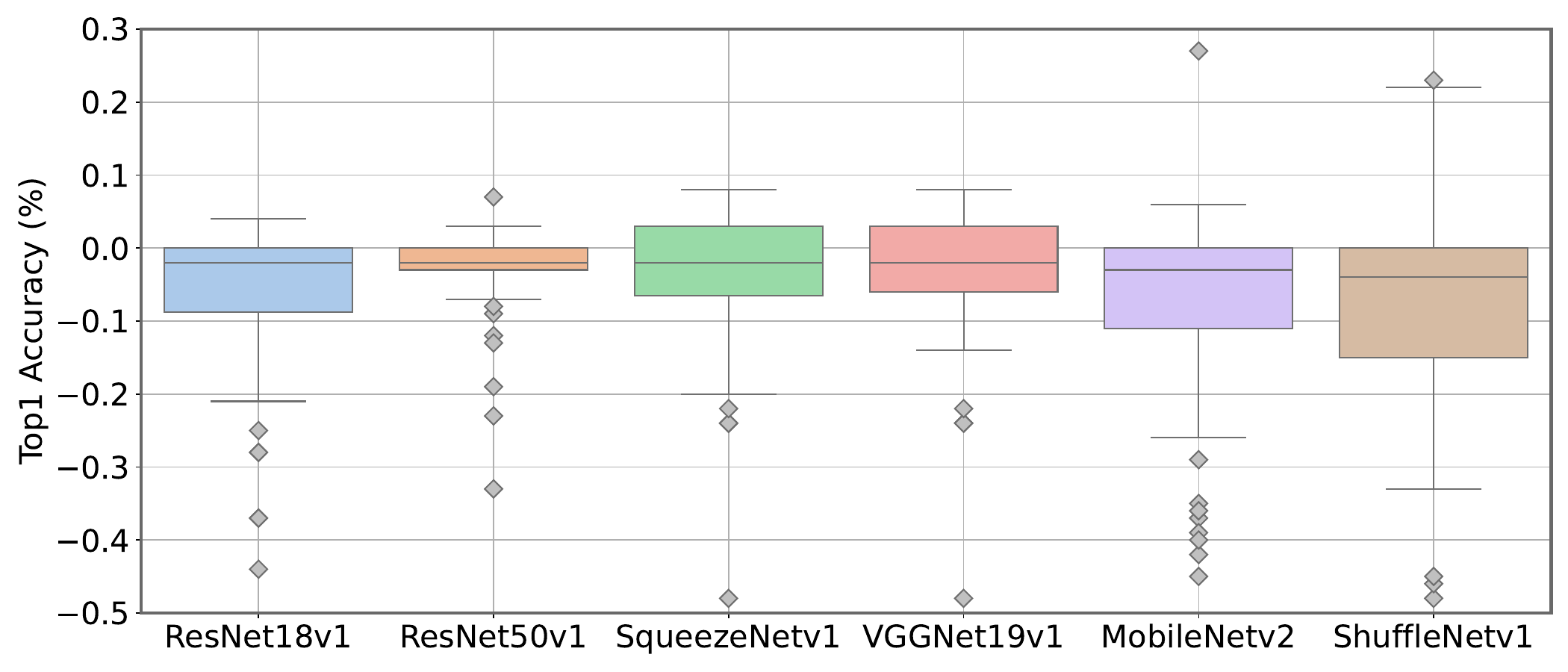}
    \caption{Layer-wise sensitivity differences in DNN models}
    \label{fig:relative_error}
\end{figure}

These methods use a metric called \textit{Sensitivity}, which relates to the model accuracy, as a criterion for assigning different precisions to each layer~\cite{ewha,wu2020integer,pandey2023practical}. \news{We measured changes in Top-1 accuracy across six DNN models as each layer was sequentially quantized to 8-bit, as shown in Fig.~\ref{fig:relative_error}. The extent of accuracy degradation varied by model.} For MobileNetv2 and ShuffleNet, some layers caused up to a 1\% decrease in accuracy even when only a single layer is quantized to 8-bit. Therefore, by identifying the layers with the highest Top-1 accuracy degradation for each model and avoiding the quantization of those specific layers, the accuracy loss can be effectively reduced.

\news{Despite the effectiveness of mixed-precision quantization, there are significant challenges when applying these methods at compile time}.
\red{As shown in Fig.~\ref{fig:runtime}, quantization aware training (QAT) methods need to be retrained, rely on the training dataset, and have complicated hyperparameter tuning, which makes it hard to use these techniques at compile time.
Mixed-precision algorithms based on Post-Training Quantization (PTQ) often encounter that the sensitivity list determined at the algorithmic stage becomes suboptimal and requires a significant amount of time to generate it.
Recent PTQ methods use reconstruction to adjust the weights, in order to reduce the loss caused by quantization. However, as shown in Fig.~\ref{fig:runtime}, this process is also quite time-consuming, making it challenging to apply at compile time.}
This is due to operator optimizations that occur in the model structure and during actual compilation process, such as fusion. 
Additionally, the weight-only-order method  ~\cite{pandey2023practical}, an existing study, ignores the characteristics of activations and relies solely on weights when generating the sensitivity list, making accurate sensitivity measurement difficult.



\news{Therefore, a lightweight mixed-precision method that can be applied at the compiler level is required, which must address three major challenges:}
\news{\textbf{C1)} Developing an efficient and simple mixed-precision method with linear computational complexity suitable for compile-time application. \textbf{C2)} Overcoming stability issues in sensitivity analysis during compile-time  by devising an effective local metric. \textbf{C3)} Selecting the optimal Graph Intermediate Representation (IR) form by considering the impact of graph IR stage selection on the overall performance of mixed-precision quantization.}

\news{
In this study, we propose {\systemName}, a solution that overcomes the three key challenges mentioned earlier by performing mixed-precision quantization during compile time based on a local metric using a small set of input data. We perform extensive experiments on five models—ResNet18v1, ResNet50v1, SqueezeNetv1, VGGNet, and MobileNetv2—focusing on model accuracy, algorithm running time, inference time, and ablation study.}

\news{The key contributions of our work are outlined below.}
\begin{itemize}


\item \red{The proposed {\systemName} has an algorithmic computational complexity of $\mathcal{O}(n)$, scaling linearly with the number of parameters in the model. This is much faster and more efficient than existing methods that need to search the exponentially large bit-width configuration space.} 

\item \news{{\systemName} can construct a stable local metric by combining weights, activations, SQNR, and MSE in compile time. Additionally, it identifies the optimal stage of the Graph Intermediate Representation (IR) for sensitivity generation by analyzing how different graph IR forms impact the sensitivity analysis and the overall performance.}

\item {\systemName} was applied to five models: ResNet18v1, ResNet50v1, SqueezeNetv1, VGGNet, and MobileNetv2 to evaluate the quantization accuracy. The results showed an average accuracy improvement of 0.66\% compared to the method using Weight SQNR~\cite{pandey2023practical}.
It achieves up to 0.68\% higher accuracy than existing methods that directly use validation datasets~\cite{wu2020integer}.

\item 
 \blue{The generation time for sensitivity lists during quantization is approximately 99.99\% faster than existing methods. Additionally, we verified {\systemName}'s execution performance on various hardware platforms, confirming that it can perform the quantization process 1.43 times faster and run quantized models up to 12.52\% faster than existing methods.}


\end{itemize}


\section{\news{Background and Related Work}} 
\news{Quantization is an important technique for optimizing deep learning models, especially for reducing model size. In this section, we introduce background knowledge and existing research on quantization, categorizing them into the areas of quantization methods and deep learning compilers. Additionally, we have highlighted the novel contributions of our work compared to existing approaches.}

\subsection{\news{Uniform Quantization and Optimization in Deep Learning Models}}
Uniform quantization for deep learning models is employed to reduce the precision of weight and activations within the neural networks. This approach aims to decrease the model's memory footprint and computational demands, which are  particularly beneficial for deploying deep learning models on devices with limited computational resources, such as mobile phones and embedded systems~\cite{app14177445}. 

Uniform quantization maps continuous real-value tensors onto a finite set of discrete values~\cite{gholami2021surveyquantizationmethodsefficient, electronics13101923}. The key characteristic of uniform quantization is that it uniformly distributes discrete values across the range of data, meaning that the intervals between adjacent quantized values are constant. This method is mathematically described as follows:

Given a real-valued input $x$, the quantized value $Q(x)$ is computed as follows:

\begin{equation}
Q(x) = \text{round}\left(\frac{x - x_{\min}}{\delta}\right)
\end{equation}

where, $x_{min}$ represents the minimum value of the tensor to be quantized, and $\delta$stands for the quantization step, which is defined as follows:

\begin{equation}
\delta = \frac{x_{\max} - x_{\min}}{N - 1}
\end{equation}

where, $x_{max}$ denotes the maximum value of the tensor, and $N$ is the number of quantization levels. The quantization levels are often chosen to be powers of two (e.g., 256 levels for 8-bit quantization), facilitating the hardware implementation of the computations.

The quantized values $Q(x)$ can then be transformed back into a dequantized representation $\hat{x}$, which approximates the original value, using:

\begin{equation}
\hat{x} = Q(x) \cdot \delta + x_{\min}
\end{equation}

This quantization process introduces a quantization error, which is defined as the difference between the original and quantized values. However, with appropriate calibration and fine-tuning, the effect of this error on the accuracy of the model can be minimized~\cite{mishra2018fixing}.

During inference, Uniform quantization is applied to the model weight and the activation (the outputs of the layers). This enables the entire inference process to be conducted using lower-precision arithmetic, significantly reducing computational complexity and power consumption.

Uniform quantization is crucial for optimizing deep learning models for edge computing. because it effectively balances the trade off between model accuracy and resource efficiency.


\subsection{Quantization}
\subsubsection{\blue{Quantization-Aware Training}}
\blue{Quantization-aware training (QAT) strategies involve adapting models from higher to lower bit precision during the training process, which is supported by numerous studies ~\cite{kris_whitepaper2018,steven2020,PACT2018,zhang2018lq,jung2019learning,zhou2016dorefa,jacob2018,songhan2016,nguyen2017variational,shin2017fixed}. For instance, Krishnamoorthi ~\cite{kris_whitepaper2018} provides foundational techniques for implementing QAT, addressing practical challenges like activation quantization and efficient inference. Jacob et al. ~\cite{jacob2018} focus on quantization and training of neural networks for efficient integer-arithmetic-only inference, which is essential for deploying models on hardware without floating-point support.
Advanced methods such as Learned Step Size Quantization (LSQ) by Esser et al. ~\cite{steven2020} and LQ-Nets by Zhang et al. ~\cite{zhang2018lq} optimize quantization scales and jointly learn network parameters and quantization functions, effectively reducing accuracy loss at low bit-widths. Choi et al. ~\cite{PACT2018} introduce PACT, where activation clipping values are learned during training, improving quantization without significant performance degradation.
The principal advantage of QAT lies in its capability to minimize the performance degradation often observed in quantized models through additional training. This approach allows for the efficient quantization of CNN architectures to lower bit representations, including 2-bit precision, without significant loss in accuracy. However, QAT faces challenges such as prolonged training times, potential unavailability of training data for third-party services, and complex hyperparameter optimization. Moreover, it is absolutely impossible to apply such methods at compile time. Attempts to address these challenges through methods like Variational Continual Learning ~\cite{nguyen2017variational} and fixed-point optimization with adaptive retraining ~\cite{shin2017fixed} have yet to fully overcome them.
Given these considerations, our study focuses on post-training quantization (PTQ), which does not require retraining and is practical for rapid deployment. PTQ is especially suitable for compiler-level applications where computational efficiency and ease of integration are paramount.} 

\subsubsection{\blue{Post-training Quantization}}

\blue{Post-Training Quantization (PTQ) converts models to lower precision without additional training, making it suitable for resource-limited devices ~\cite{jiang2021automated,banner_neurips2019,choukroun2019low,nagel2019data,zhao2019improving,lee2018quantization,goncharenko2019fast,meller19a,migacz20178,wu2020integer,novak2018sensitivity,Lee_2022,adaquant,brecq21,wei2022qdrop}. It addresses the time consumption and data scarcity concerns associated with QAT but can lead to significant accuracy loss, especially at lower bit-widths.
To enhance PTQ performance, Wu et al. ~\cite{wu2020integer} provided principles and empirical evaluations for integer quantization. Nagel et al. ~\cite{nagel2019data} proposed data-free quantization using weight equalization and bias correction, mitigating accuracy degradation without training data. Banner et al. ~\cite{banner_neurips2019} introduced post-training 4-bit quantization with minimal accuracy loss, while Zhao et al. ~\cite{zhao2019improving} improved quantization without retraining by addressing outlier channels.
AdaQuant by Hubara et al. ~\cite{adaquant} minimizes layer-wise quantization errors post-training through iterative optimization and integer programming, using unlabeled data to calibrate activations without requiring fine-tuning. BRECQ by Yuhang Li et al. ~\cite{brecq21} minimizes second-order quantization errors using gradient-based optimization involving complex computations like Hessian matrices. QDrop by Xiuying Wei et al. ~\cite{wei2022qdrop} introduces quantization-aware dropout to mitigate quantization noise. However, these methods require iterative training steps and weight updates.
\news{QuantuneV1~\cite{Lee_2022} proposed a method for exploring various quantization configurations and recommending the optimal approach for each model. However, it only provided an ON/OFF functionality for mixed precision and did not explore the optimal mixed precision for all layers.} Despite these advancements, PTQ methods often struggle to maintain consistent accuracy across different models due to architectural variations and data distributions. Determining optimal quantization settings for each model can be time-consuming and lacks a universal solution. We focused our research on quickly identifying effective parameters for each model that can be applied at compile time to maintain high accuracy within a short time frame. }

\subsubsection{Mixed-Precision Quantization}
Prior mixed-precision works formulated the mixed-precision optimization problem through reinforcement learning to allocate the optimal bit precision per layer of a deep neural network. 
In such approaches, a reinforcement learning agent determines the bit width of each layer and learns to determine optimal balance between the model performance and cost reduction through the rewards associated with these decisions~\cite{guo2018survey}. The key to this method is automating the complex bit allocation process and allowing the agent to efficiently explore the search space through various reinforcement learning algorithms~\cite{gholami2021survey}. A specific example is the HAQ framework which  hardware feedback is incorporated to learn policies that minimize energy and latency while maintaining accuracy~\cite{wang2019haq}. This reinforcement learning-based optimization method is particularly useful for reducing computational costs while maintaining model performance, but it takes a long time.
Also, these methods have limitations. First, determining bit-widths through sensitivity analysis is complex. Second, they rely heavily on training data, limiting generalization to new domains. Third, these methods can still incur high computational costs and long execution times.


\subsection{\purple{Deep Learning Compilers}}
The surge in efficiency requirements for deep learning models has underscored the importance of DL compilers, which have been developed across academia and industry. Although proprietary compilers, such as Intel nGraph ~\cite{cyphers2018intel}, NVIDIA TensorRT ~\cite{migacz20178}, ARM NN, and Xilinx Vitis focus on specific hardware platforms, open-source alternatives such as TVM ~\cite{chen2018tvm} and Glow ~\cite{rotem2018glow} offer broader hardware support. However, these community-driven compilers typically require manual exploration to determine the optimal quantization settings, which can delay model deployment. \purple{Manual exploration becomes impractical due to the exponentially increasing search space as the number of layers grows ~\cite{dong2019hawq}. For example, a model with 100 layers and two possible bit-widths per layer results in $2^{100}$ possible configurations, which is infeasible to explore exhaustively. This issue leads to significant time and resource consumption, especially for larger models or when deploying across different hardware platforms.}
In this study, {\systemName}, enhances these compilers by introducing an advanced mixed-precision algorithm capable of automatically determining the best configuration for accuracy. \purple{By efficiently determining appropriate bit-widths at compile time using sensitivity metrics like MSE and SQNR, {\systemName} eliminates the need for manual exploration. }



\section{\red{Detailed Challenges of Applying Mixed-Precision Quantization at Compile Time}}
\label{sec:chall}
In this section, we outline three significant challenges (C1–C3) that hinder the effective application of mixed-precision quantization at compile time.

\subsection{C1: \red{Demand for Efficient and Simple Mixed-Precision Methods}}
\label{subsec:c1}
The first challenge is efficiently determining mixed-precision quantization during compile time. To apply mixed precision at the compiler level in a single optimization pass, the process needs to be as fast and simple as possible.
However, complex methods that require real-time weight updates or error reconstruction are inefficient and impractical. In other words, existing PTQ methods based on reconstruction and QAT that require backpropagation cannot be applied to compile-time quantization due to computational complexity.
\red{Additionally, leading frameworks for optimizing deep learning models such as TVM and Glow do not support mixed-precision quantization at compile time. Therefore, a fast and simple mixed-precision quantization method that can be executed within compile time is needed.}


\begin{figure}[t]
    \centering
    \subfloat{\label{fig:ResNet18_sqnr_graph}\includegraphics[width=1\linewidth]{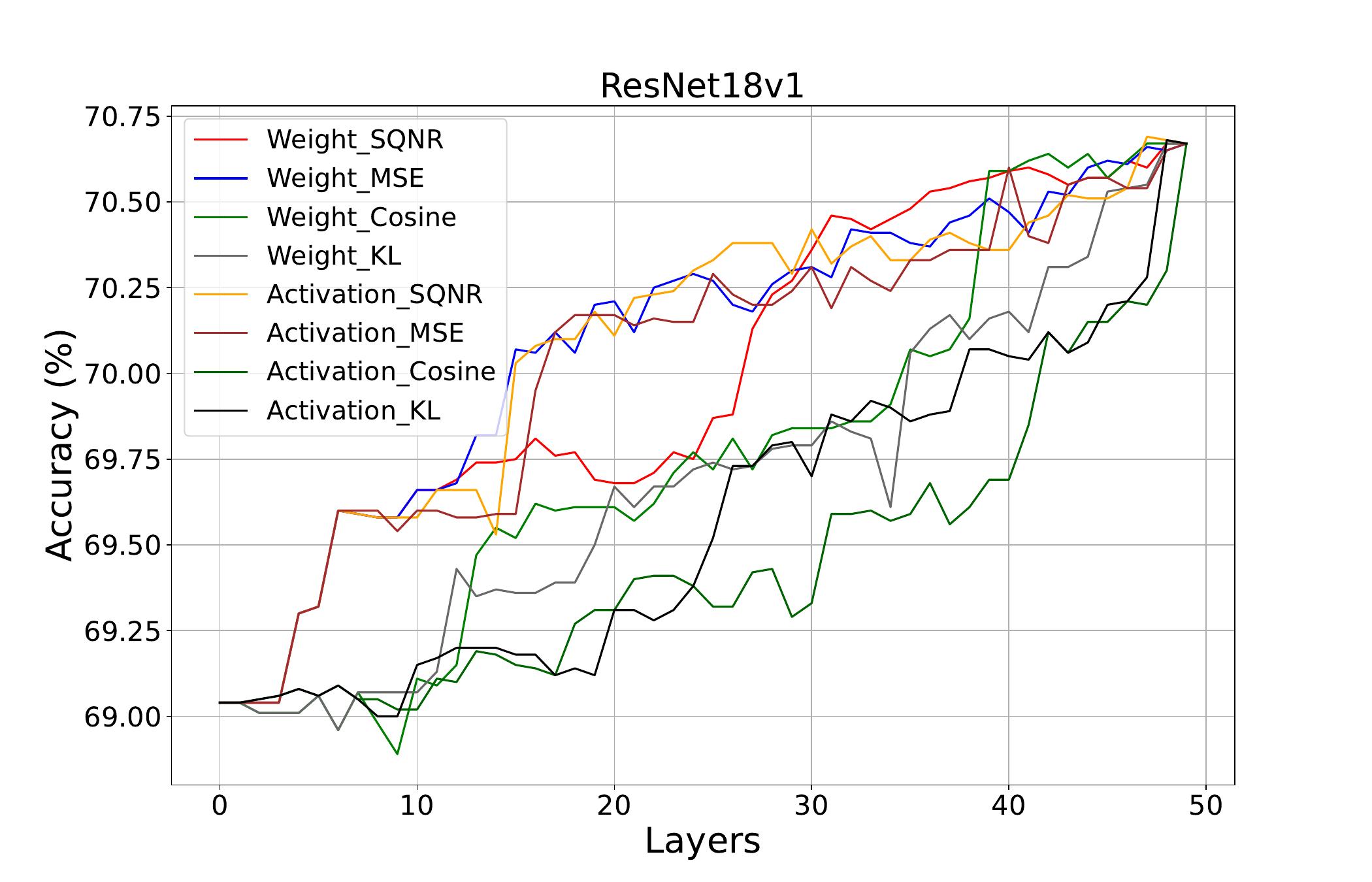}} \vfill
    \subfloat{\label{fig:mobile_local_graph}\includegraphics[width=1\linewidth]{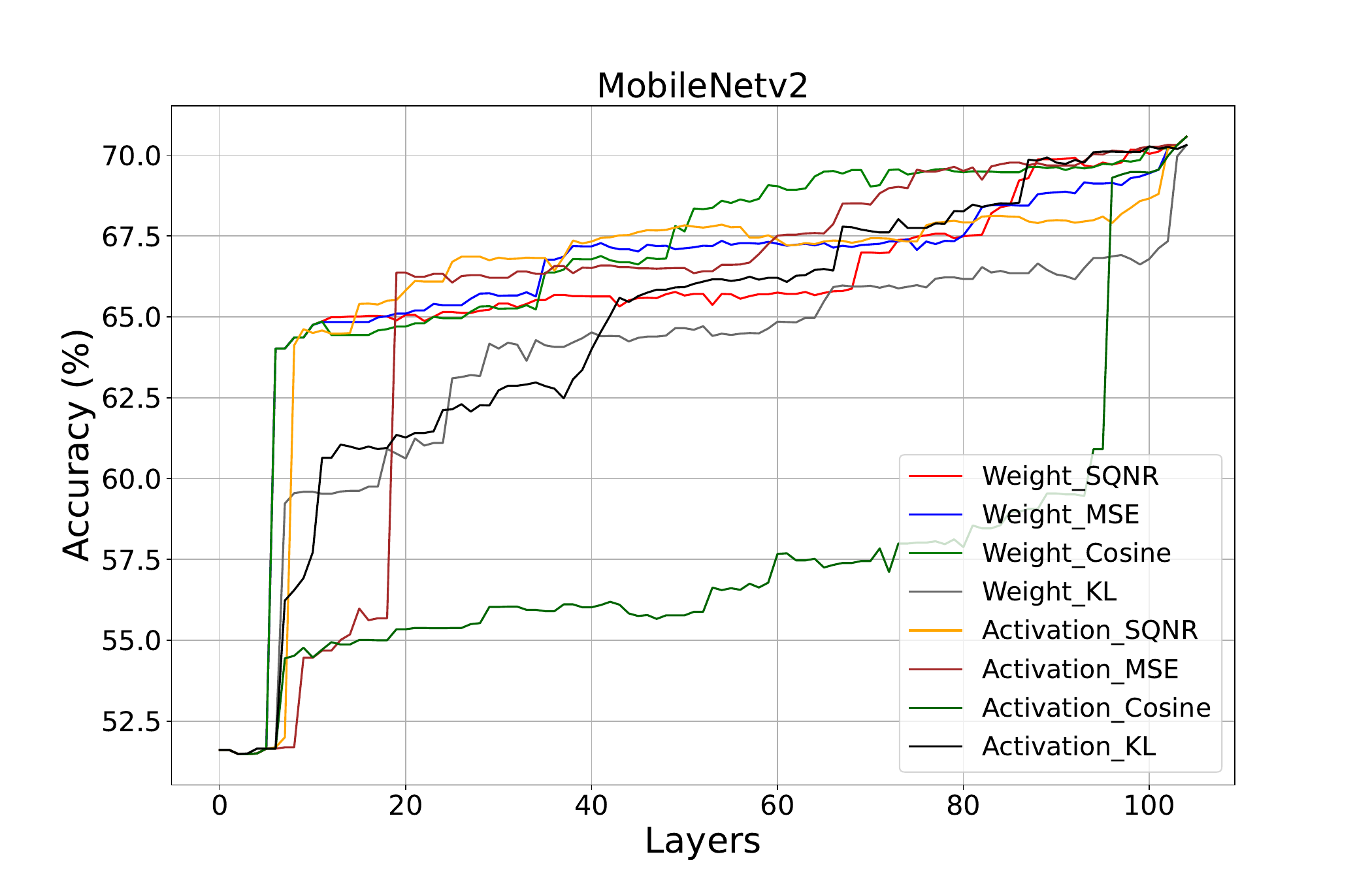}}
    \caption{\red{Results of applying mixed-precision to ResNet18v1 and MobileNetv2 across six local metrics (performing dequantization one layer at a time)}}
    \label{fig:local}
\end{figure}

\subsection{C2: \red{Stability Issues in Sensitivity Analysis}}
\label{subsec:c2}

The second challenge is \red{measuring and ranking the quantization sensitivity of each layer in a DNN model} for mixed-precision quantization~\cite{jacob2017,ewha,wu2020integer,pandey2023practical}. It is necessary to determine a local metric with strong generalization performance to accurately measure the sensitivity of each layer. Existing studies propose sensitivity measurement methods, such as the Jacobian Norm~\cite{zhang2017understanding} or First-Order Sensitivity~\cite{Chauhan_2023_ICCV}; however, they are limited by computational complexity and inconsistent results depending on the input images. 
\red{In this study, we consider SQNR ~\cite{lin2016fixedpointquantizationdeep}, MSE ~\cite{banner2019posttraining4bitquantizationconvolution,choukroun2019lowbitquantizationneuralnetworks}, and cosine similarity \cite{Angularquantization}, KL divergence~\cite{Xu_2021,pmlr-v222-matsumoto24a} as applicable local metrics for application during compile time. SQNR has been used as a metric for evaluating quantization noise in neural networks. MSE is used to measure quantization error before and after quantization, and cosine similarity is employed to assess the similarity between original and quantized weights. KL divergence is a statistical measure used to evaluate the difference between two distributions before and after quantization. These metrics are computationally efficient and can be applied to both weights and activations, resulting in eight combinations.}

\red{Fig.~\ref{fig:local} shows the results of sequentially applying dequantization to ResNet18v1 and MobileNetv2 using the eight combinations of local metrics. Accuracy recovery varied depending on the local metric used. Quantization using SQNR and MSE showed similar overall accuracy recovery performance. In contrast, cosine similarity and KL divergence showed significant variations in accuracy recovery performance depending on weights and activations. }

\red{Therefore, it is crucial to select a local metric with strong generalization performance for consistent application to DNN models. Additionally, it is necessary to analyze whether to use random images or images used for calibration, and determine which values among weight and activation tensors based on input images are most effective.}


\begin{figure}[t]
    \centering
    \includegraphics[width=0.85\linewidth]{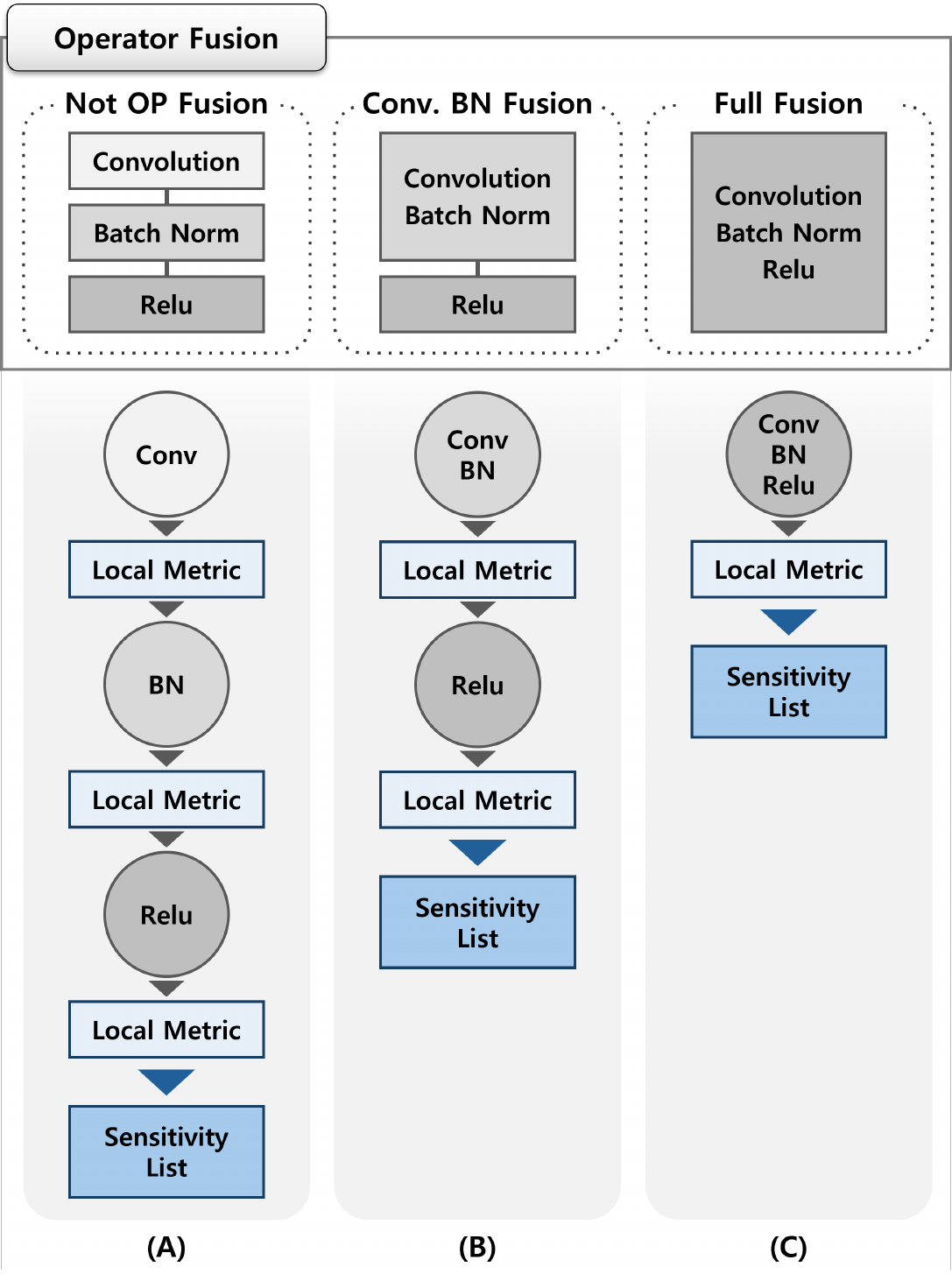}
    \caption{\blue{Generation of local metrics according to the stages of operator fusion}}
    \label{fig:fusion}
\end{figure}

\subsection{C3: \news{Selection of Optimal Graph Intermediate Representation}}
\label{subsec:c3}
The third challenge is determining the appropriate stage of the Graph Intermediate Representation (Graph IR) at which to calculate the local metric for generating the sensitivity list during compile time to achieve optimal performance. Graph IR is an abstract representation used by compilers to represent the intermediate stage of a program, allowing for various optimizations and transformations. \news{As shown in Fig.~\ref{fig:fusion}, there are multiple levels of operator fusion that can be applied. These include no operator fusion (A), partial fusion of Convolution and Batch Normalization (B), and full fusion of Convolution, Batch Normalization, and ReLU (C).} At the Graph IR level, the final Graph form is modified as in (C) to reduce operation execution time and memory access overhead by executing multiple operations in a single kernel~\cite{deep_learning_inference_optimisation,Acharya_2020}. \news{However, if the sensitivity list is generated from a fully fused Graph IR (C), it becomes difficult to accurately evaluate the sensitivity of individual operators because the inherent sensitivity of multiple fused operators is represented as a single sensitivity.}

\news{As shown in Table~\ref{table:Graph IR}, the sensitivity list varies depending on the operator's fusion stage, affecting the model's accuracy. Additionally, the number of operations due to quantization, de-quantization iterations changes, resulting in variations in the overall execution speed of the model. Therefore, it is necessary to analyze how generating a sensitivity list and computing the local metric at different Graph IR levels affect quantization performance. First, it is necessary to consider methods that calculate operator-specific local metrics at the non-fused stage (A), and then apply operator fusion at stage (C) to speed up operator execution.}


\begin{table}[ht]
\centering
\caption{Results of mixed-precision quantization accuracy improvement at each IR stage of the graph (A: not operator fusion, A-Fusion : after A, OP fusion, C: full OP fusion, \# of Q-DQ: number of quantization and dequantization operations with 40\% quantization applied)}
\resizebox{\columnwidth}{!}{
\label{table:Graph IR}
\begin{tabular}{@{}lccccccc@{}}  
\toprule
  \multirow{3}{*}{\ \ \ \ \ \ \ Model}  & \multicolumn{2}{c}{\textbf{A}} & \multicolumn{2}{c}{\textbf{A-C (Ours)}} & \multicolumn{2}{c}{\textbf{C}}  \\ \cmidrule(l){2-7} 
     & Acc & \# of Q-DQ & Acc & \# of Q-DQ & Acc & \# of Q-DQ \\ \midrule
     \text{\ ResNet18v1}  & 70.17\% &9  & 70.31\% &7  & 69.84\% &7 \\
     \text{\ ResNet50v1}  & 75.12\% &37 & 75.12\% &35 & 74.61\% &33 \\
     \text{MobileNetv2}   & 66.62\% &37 & 66.88\% &32 & 66.44\% &28 \\ 
\bottomrule
\end{tabular}
}
\end{table}

\section{Overview}

\begin{figure*}
    \centering
    \includegraphics[width=0.85\linewidth]{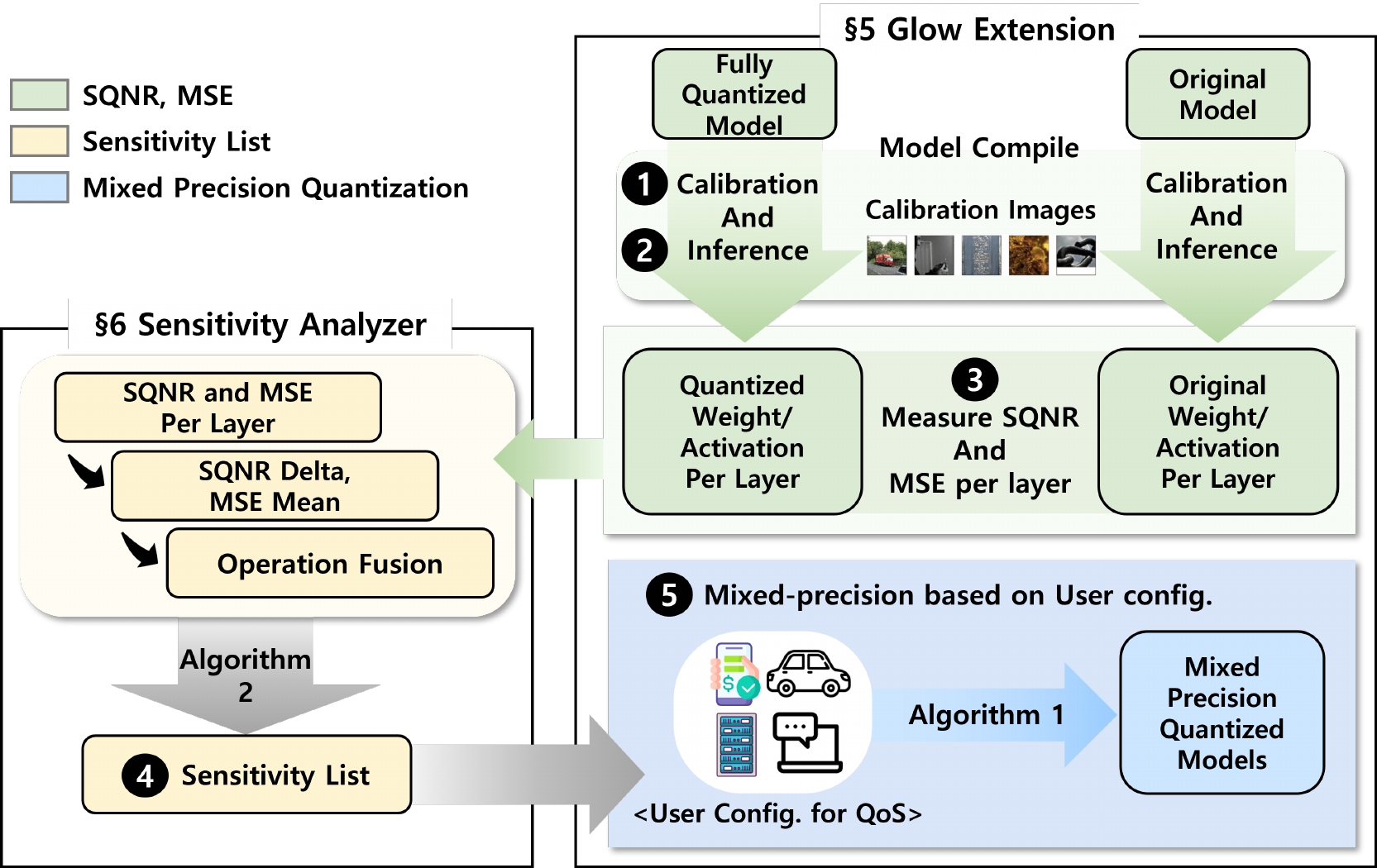}
    \caption{\blue{Overview of {\systemName}}}
    \label{fig:overview}
\end{figure*}

The proposed mixed-precision method extends the capabilities of DNN compilers to support mixed-precision quantization at the compiler level. This approach quickly identifies layers in the input model that can significantly degrade accuracy during compile time and applies mixed-precision accordingly. To overcome the three challenges mentioned in \S\ref{sec:chall}, the proposed {\systemName} employs three strategies: \purple{i) By performing inference only twice—before and after quantization—for the entire model, we achieve a computational complexity of $\mathcal{O}(n)$ that increases linearly with the number of parameters; ii) It uses a stable local metric that combines weights, activations, SQNR, and MSE to determine mixed-precision without relying on backpropagation; and iii) Select the most appropriate intermediate representation (IR) from the various IRs generated during the compilation process to calculate layer-wise sensitivity. Subsequently, to enhance operational execution speed, perform operator fusion into the final graph form, thereby ensuring both high model accuracy and execution speed.} From the user's perspective, mixed-precision is determined by specifying the desired level of quantization according to their objectives.

Specifically, the operation of {\systemName} is divided into two main parts: sensitivity analyzer and a model compiler, as shown in Fig.~\ref{fig:overview}. The execution process follows five sequential steps:
\blackcircle{1} \purple{Data Distribution Verification and Calibration: Collect statistical data on the distribution of weights and activations by running the model on a small calibration dataset. Perform calibration to adjust model parameters and ensure that the quantization scales accurately represent the data range of each layer.
\blackcircle{2} Scale Information Calculation and Initial Quantization: Based on the collected data distribution, calculate the scale information for quantization of the weights. For symmetric quantization, the scale (S) for each layer's weights is calculated using:
\begin{equation}
S = \frac{2^{b - 1} - 1}{\max(|W|)}
\end{equation}
where W represents the weight tensor, and b is the bit-width. The quantization process maps the floating-point weights to integer values using this scale:
\begin{equation}
Q(W) = \text{round}\left( \frac{W}{S} \right)
\end{equation}
We then apply uniform quantization across all layers with the same bit-width to create an initial quantized model.}

\blackcircle{3} \purple{Local Metric Computation (SQNR and MSE Measurement): Measure the SQNR and MSE of each layer by comparing the weights and activations of the original and quantized models. This provides quantitative metrics for the quantization error and helps identify layers that are sensitive to quantization.}
\blackcircle{4} \purple{Sensitivity Analysis and List Generation: Calculate the SQNR delta for each layer, which is the difference in SQNR between adjacent layers. Generate a sensitivity list for quantization by combining the SQNR delta, MSE values, and considering the effects of operator fusion (e.g., Convolution and ReLU).}
\blackcircle{5} \purple{Configuration File Generation and Mixed-Precision Application: Generate a configuration file that specifies the bit-width for each layer according to the desired quality of service (QoS) requirements. Using the quantization function of the Glow Compiler\footnote{\texttt{transformForPrecisionMode()} optimization pass was modified to support mixed-precision mode.}, layers identified as sensitive maintain 32-bit precision, while the remaining layers are quantized to 8-bit precision (\S~\ref{subsec:glow_extension}).}

\section{Compiler Extension for Mixed-Precision Quantization}
\label{subsec:glow_extension}

This section outlines a comprehensive methodology for the quantization and design of each component involved. 
The proposed mixed-precision quantization method can be extended and applied to all DNN compilers. In this paper, we extended the Glow compiler to create {\systemName} in order to validate the performance of mixed-precision quantization at the compiler level.
Therefore, it supports compiler backends on various hardware platforms.
With respect to Glow, our extension is designed to accommodate integer-only accelerators, and enabled layer-wise quantization for mixed-precision. The quantization process comprised two stages: calibration and mixed-precision.

\textbf{Calibration stage.} During this phase, we captured a histogram of potential numeric ranges for the activation of each neural network layer and stored it in a calibration cache. For initial calibration purposes, the Glow compiler processes a pre-trained model and 100 randomly selected images from the training set as input ~\cite{Lee_2022}. By monitoring the execution during inference, we generated a histogram of tensor values to identify the quantization errors in the activations and weights across each neural network layer.

\textbf{Mixed-Precision Decision Stage.}
The application of mixed-precision to each layer using the list determined by the Sensitivity Analyzer is a critical stage. The Glow Compiler fundamentally offers setting\footnote{\texttt{keep-original-precision-for-nodes}} to compensate for accuracy degradation due to quantization. These settings allow all operator types to remain unquantized. Although this is a straightforward method, it does not permit the precise selection of layers for quantization, leading to limitations in its use. For instance, for the ResNet18v1 neural network model, the convolution operators constituted 20 of the total 48 layers. Not quantizing all convolution layers is a burdensome approach when mixed-precision is used. Therefore, the features provided by the default Glow Compiler impose a significant constraints on its usage. 

In this study, to address these constraints, we extended the functionality of the Glow Compiler to support the selection of quantization applications on a per-layer basis. The total search space for the mixed-precision considered in this study was calculated as follows: Mixed-precision \( B \) was determined to be 2 using FP32 and INT8. If the total number of layers is \( L \), the search space is \( B^{L} \). Considering all dependencies, the search space becomes \( 2^{48} \) for ResNet18v1, making it non-tractable. Therefore, as in previous mixed-precision studies, this research also assumes that each layer is independent of the others~\cite{dong2019hawq,dong2020hawq,yao2021hawq}. In this case, the search space of \( B^{L} \) simplified to \( BL \) with a  linear complexity, making it a tractable problem.

The specific method for applying mixed-precision to the Graph IR level generated by the compiler is shown in Algorithm~\ref{mixed-algo}.
\texttt{transformForPrecisionMode()} involves specific modifications within the Glow Compiler to optimize performance and efficiency.
 The inputs of Algorithm~\ref{mixed-algo} are a DAG Graph \( F \) and a dequantized node list \( L \). The dequantized node list \( L \) consists of the names of the layers that will not be quantized, as determined by the Sensitivity Analyzer from the Sensitivity List, according to the user-defined level of quantization. Therefore, quantization is not applied to sequentially set the layers listed in the dequantized node list \( L \). The final output is the Quantized DAG Graph \( F^{*} \). As shown in lines 1 and 2, pointers corresponding to the start and end nodes are allocated from the input graph \( F \) and stored in nodeIt and stopIt. As described in lines 3--22, starting from the last node, if the node is not stored in \( L \), it is converted into a quantized node, and the existing graph node is replaced. After converting the graph nodes to quantized ones, basic compiler optimizations such as Dead Code Elimination (DCE) and Common Subexpression Elimination (CSE) are performed as post-processing to remove unnecessary parts from the existing graphs. The specific part matching \( L \) is detailed in Lines 8--14.

\begin{algorithm}[t]
  \caption{Graph node level mixed-precision quantization}
  \label{mixed-algo}
  \hspace*{\algorithmicindent}\textbf{Input:} DAG Graph $F$ \\
  \hspace*{\algorithmicindent}\textbf{Input:} Dequantized node list $L$ \\
  \hspace*{\algorithmicindent}\textbf{Output:} Quantized DAG Graph $F^{*}$ 
  
  \begin{algorithmic}[1]
  
    \LState $nodeIt \gets F.getNodes().end()$ \textcolor{blue}{\Comment{Initialize start node}}
    \LState $stopIt \gets F.getNodes().begin()$ \textcolor{blue}{\Comment{Initialize end node}}
    \Repeat
        \LState $nodeIt \gets nodeIt - 1$
        \LState $node \gets *nodeIt$
        \LState $node\_name \gets node.getName()$
        \LState $isExist \gets false$
        \While{$L$ is not the end of list}
            \LState $inStr \gets$ $L$ \textcolor{blue}{\Comment{Read a line from the sensitivity list}}
            \If{$inStr$ is $node\_name$}
                \LState $isExist \gets true$
                \LState Break
            \EndIf
        \EndWhile
        \If{$isExist$}
            \LState Continue
        \EndIf
        \LState $convertOutputs(node)$ \textcolor{blue}{\Comment{Output scales for quantization}}
        \LState $convertInputs(node)$ \textcolor{blue}{\Comment{Input scales for quantization}}
        \LState $newNode \gets newNode(node)$ \textcolor{blue}{\Comment{Replace the node}}
        \LState $postProcessing(newNode)$ \textcolor{blue}{\Comment{DCE, CSE, etc}}
    \Until{$nodeIt = stopIt$}
    \LState $cleanUp()$
  \LState \Return $F^{*}$
  \end{algorithmic}
\end{algorithm}

\section{Sensitivity Analyzer}
\label{sec:sensitivity_analyzer}
To establish a reliable sensitivity list for evaluating mixed-precision quantization, the Sensitivity Analyzer considers the following factors: 
i) careful selection of appropriate local metrics, ii) exploration of suitable Graph IR and calibration techniques, iii) optimization of quantization during operator fusion, and iv) utilization of the SQNR delta.
Each design choice is explained in detail below.

\subsection{Selection of local metrics}
\label{subsec:local_metric}
In the sensitivity analyzer, to apply mixed-precision quantization, a sensitivity list was generated using the local metrics of  SQNR and MSE simultaneously. 
As mentioned in Section~\ref{subsec:c2}(C2), the available local metrics at compile time include SQNR, MSE, and cosine similarity, KL divergence. 
To compute the local metrics, the value outputs for each layer before and after quantization were stored during the compilation phase. 
The local metrics considered in this study are SQNR, MSE, cosine similarity and KL divergence, each with specific calculation methods as described below.
\textbf{SQNR} measures the ratio of signal power to quantization noise power, where higher values indicate lower quantization noise~\cite{SQNR}. 
The actual calculation follows Eq.~\eqref{eq:sqnr}, focusing on the overall energy ratios, and thereby potentially ignoring small errors in specific regions.
It is particularly useful in identifying regions with significant quantization noise, as it provides an overall measure of signal quality.

\begin{equation}
\text{SQNR} = 10 \cdot \log_{10} \left( \frac{P_{\text{signal}}}{P_{\text{noise}}} \right)
\label{eq:sqnr}
\end{equation}

The \textbf{MSE} computes the average of the squared differences between the predicted and actual values. 
The calculations are shown, in Eq.~\eqref{eq:mse}, indicates that lower values correspond to fewer quantization error\cite{choukroun2019lowbit}. 
The MSE focuses on the absolute differences between the actual and predicted values, with the squared terms emphasizing larger errors.
\begin{equation}
\text{MSE} = \frac{1}{n} \sum_{i=1}^{n} (P_{\text{signal}} - P_{\text{noise}})^2
\label{eq:mse}
\end{equation}

\purple{\textbf{Cosine Similarity} measures the angle between two vectors to compute similarity~\cite{Gong2012AngularQB}. 
The calculation follows Eq.~\eqref{eq:cosine}, which focuses on the vector direction without considering its magnitude. 
Values range between -1 and 1, where proximity to 1 signifies similar vector directions. Cosine similarity does not consider the signal magnitude or energy levels and potentially disregards information about the overall signal strength.
While the angles may be similar, significant differences in the actual values can exist, thus limiting their ability to evaluate the  absolute quantization error size. Moreover, applying it across various DNN models may pose challenges in ensuring consistent performance.}
\begin{equation}
\text{cosine similarity}(\mathbf{A}, \mathbf{B}) = \frac{\mathbf{A} \cdot \mathbf{B}}{\|\mathbf{A}\| \|\mathbf{B}\|}
\label{eq:cosine}
\end{equation}
\news{\textbf{KL Divergence} measures the difference between two probability distributions \( P \) (original layer output) and \( Q \) (quantized layer output)~\cite{pmlr-v222-matsumoto24a}. It provides insight into how much information is lost when converting high-precision activations and weights to a lower bit-width format. KL Divergence is defined as:}

\begin{equation}
\news{\text{KL Divergence}(P \parallel Q) = \sum_x P(x) \log \left( \frac{P(x)}{Q(x)} \right)}
\label{eq:kld}
\end{equation}

\news{A lower KL Divergence indicates that the quantized output closely matches the original, signaling a successful preservation of information during quantization. Higher values of KL Divergence suggest greater deviations, indicating layers that are more susceptible to accuracy loss after quantization. However, when applying KL Divergence to various DNN models, sensitivity analysis is not properly performed, so it has clear limitations as a local metric for mixed-precision quantization.}


Therefore, this study employs a strategy that combines SQNR and MSE as complementary local metrics. The concurrent use of SQNR and MSE is motivated by their synergistic effects, as illustrated in Fig.~\ref{fig:Gradient of SQNR, MSE for Resnet18} for ResNet18v1 experiments, showing that they exhibit similar trends while complementing each other. When significant quantization errors occur, it can be observed that both weight and activation SQNR values decrease, while the MSE values increase significantly. Thus, SQNR evaluates the overall signal-to-noise ratio, while MSE assesses quantization errors at the individual sample level. Combining these two local metrics enables a more comprehensive and detailed analysis.

\begin{figure}[t]
	\centering
\subfloat{\label{fig:res18_sqnr_graph}\includegraphics[width=1\columnwidth]{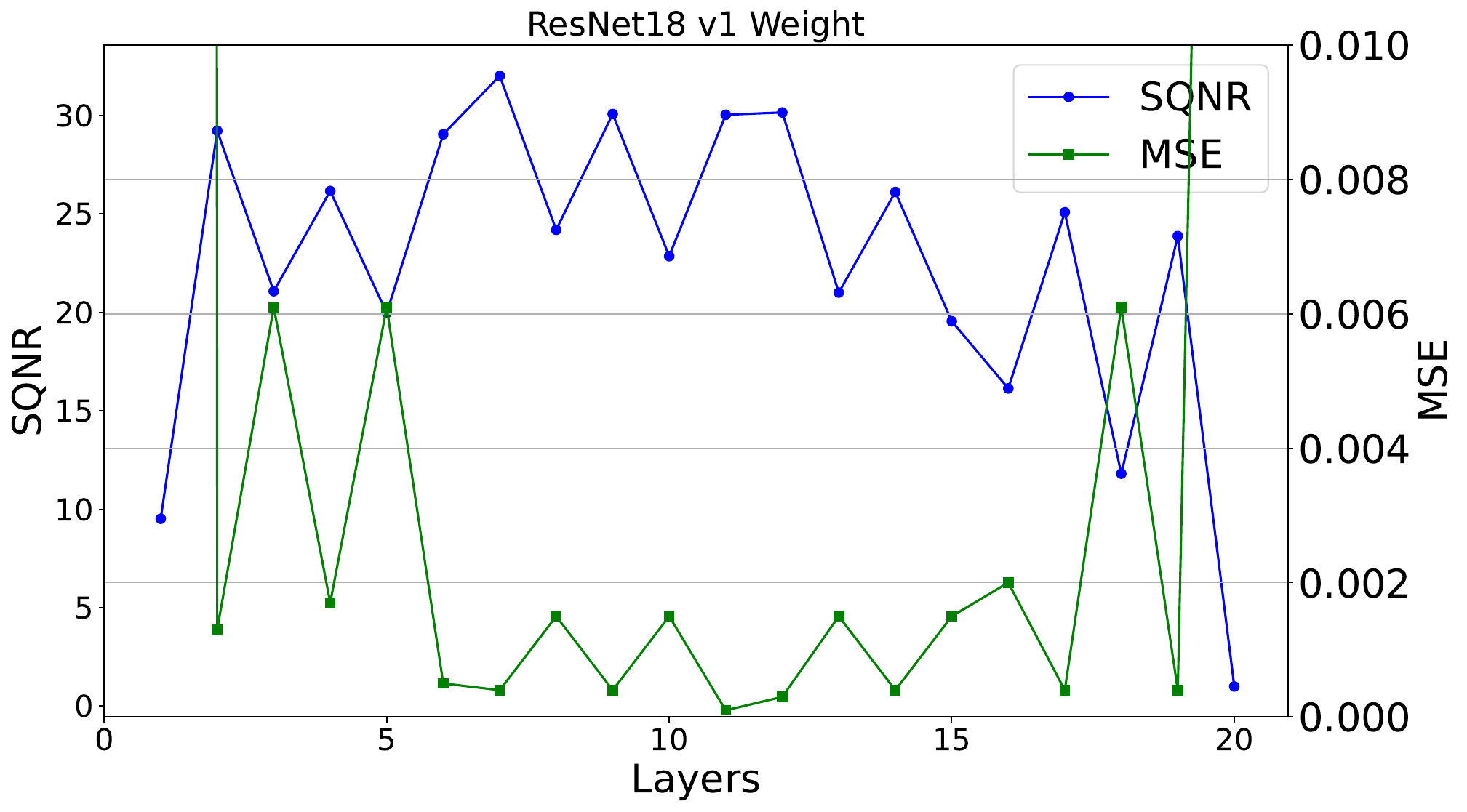}} \vfill
\subfloat{\label{fig:Mobile_sqnr_graph}\includegraphics[width=1\columnwidth]{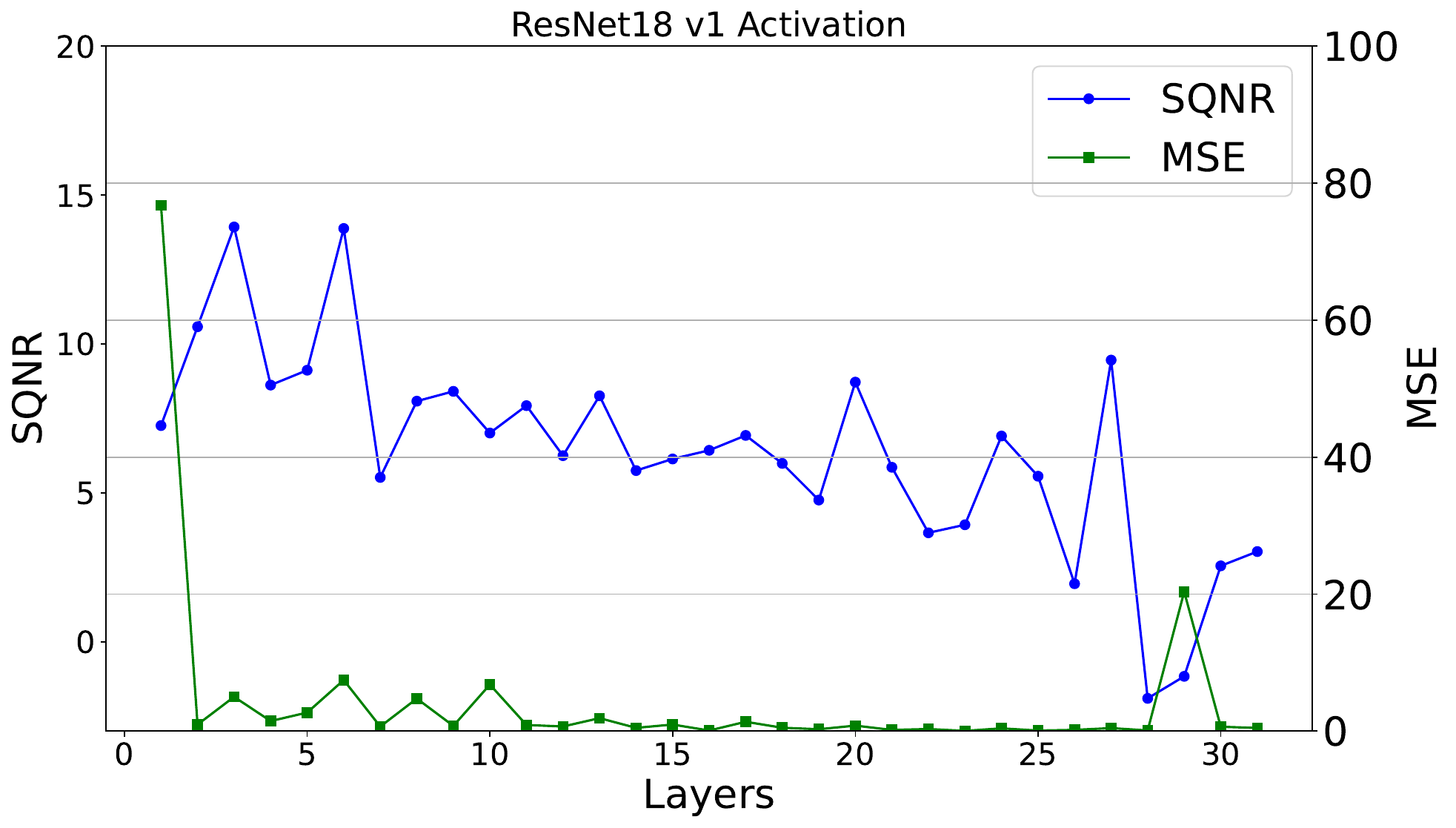}}
    \caption{Changes in weight, activation SQNR, and MSE for each layer of ResNet18v1}
    \label{fig:Gradient of SQNR, MSE for Resnet18}
\end{figure}

\subsection{Optimizing Graph IR Search and Calibration for Optimal SQNR Measurement}
\label{subsec:impact_of_operator_fusion}
\begin{figure}[t]
    \centering
    \includegraphics[width=1\linewidth]{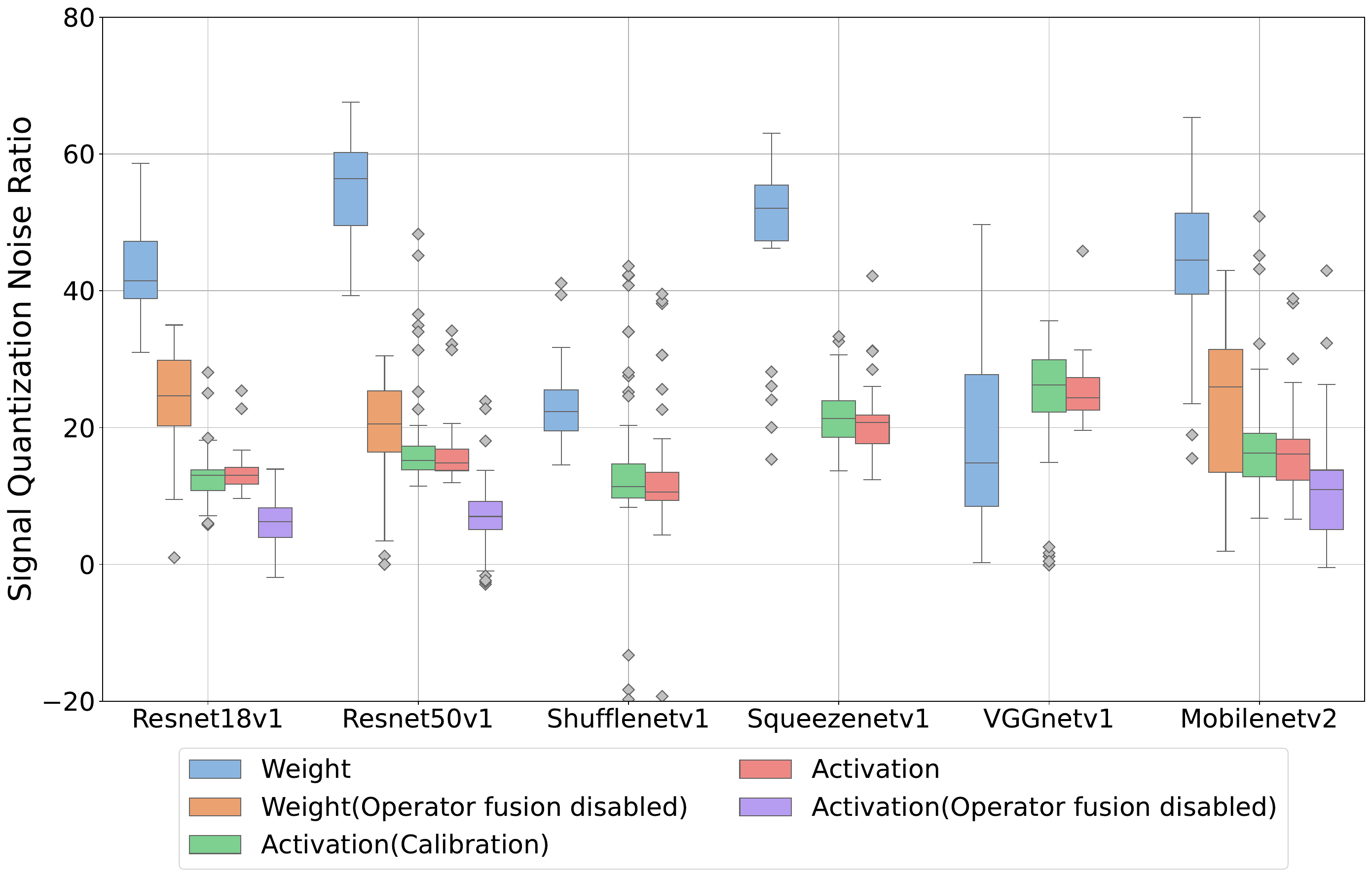}
    \caption{Range of SQNR values for W8A8 quantizers in different networks}
    \label{fig:Boxplot}
\end{figure}

Experiments were conducted to determine how to calculate the optimal SQNR for the weights and activations.
This exploration was divided into two parts: one focusing on operator fusion, as mentioned in Section~\ref{subsec:c3} (C3), and the other on calculating activation tensor. 
Fig.~\ref{fig:Boxplot} illustrates the distribution of SQNR values when these two aspects are considered in the six models.
As shown in Fig.~\ref{fig:Boxplot}, the distribution of SQNR varies among models, as previously reported\cite{pandey2023practical}. 
Furthermore, it is evident that operator fusion significantly affects the SQNR values for both weight and activation. 
In general, the SQNR values for weights showed a wider range when operator fusion was not applied compared to when it was applied. By disabling the operator fusion, the SQNR can be calculated for each convolution and batch normalization operator individually, which better reflects the unique characteristics of each layer and more effectively distinguishes the sensitivity differences between layers. In addition, the expanded precision options allocated individually to each operator increase the search space, enabling optimal precision allocation for each layer. Therefore, in this study, sensitivity list generation was performed without operator fusion.

Another important factor is that the Signal-to-Quantization Noise Ratio (SQNR) for activation varies depending on the input image. Therefore, we analyzed the SQNR values calculated using both calibration and arbitrary datasets. As the weight distribution of the DNN model was tailored to the characteristics of the calibration dataset, obtaining the activation SQNR with the calibration dataset allowed the SQNR to be distributed over a wider range. This makes it better at distinguishing sensitivity differences between layers. Therefore, calculating the SQNR using the calibration dataset is more advantageous for creating sensitivity lists.

\subsection{Operator Fusion Optimization for Optimal Mixed-Precision}
\label{subsec:operator_fusion}
In this study, fusion was not applied during the sensitivity analysis of the layers. However, a fused graph IR was used when applying mixed-precision. This is because, as explained earlier, the sensitivity list considers the form of layers before fusion for precision analysis, Whereas at the application stage, the fusion of layers is considered for accuracy and execution speed optimization. The considered fusion involves batch normalization and the ReLU activation function as follows.

\purple{Eq.~\eqref{eq:conv_batch_seq} illustrates the computation of the output $Y_i$ at the $i$-th layer when a convolution operation is followed by batch normalization. The convolution $\text{conv}(x)_i$ is defined as $\text{conv}(x)_i = W_i x_i + B_i$, where $W_i$ and $B_i$ represent the weights and biases of the convolutional layer, respectively.}

\purple{\begin{equation} Y_i = \gamma_i \left( \frac{\text{conv}(x)_i - \mu_i}{\sqrt{\sigma_i^2}} \right) + \beta_i, \quad \text{conv}(x)_i = W_i x_i + B_i 
\label{eq:conv_batch_seq} 
\end{equation}}

\purple{By substituting the convolution expression into the batch normalization formula, we obtain the fused operator in Eq.~\eqref{eq:fusion_conv_batch}:}

\purple{
\begin{align} 
Y_i &= \gamma_i \left( \frac{W_i x_i + B_i - \mu_i}{\sqrt{\sigma_i^2}} \right) + \beta_i \nonumber \ \\
&= \left( \frac{\gamma_i}{\sqrt{\sigma_i^2}} W_i \right) x_i + \left( \frac{\gamma_i (B_i - \mu_i)}{\sqrt{\sigma_i^2}} + \beta_i \right) \nonumber \ \\ 
&= A_i x_i + C_i 
\label{eq:fusion_conv_batch} 
\end{align}
}

\purple{The fused weights and biases are defined in Eq.~\eqref{eq:fused_w_b}.}

\purple{
\begin{align} 
A_i &= \frac{\gamma_i}{\sqrt{\sigma_i^2}} W_i \ C_i &= \frac{\gamma_i (B_i - \mu_i)}{\sqrt{\sigma_i^2}} + \beta_i 
\label{eq:fused_w_b} 
\end{align}
}

\purple{This fusion integrates batch normalization into the convolution operation by adjusting the original weights and biases to $A_i$ and $C_i$. By precomputing these adjusted parameters during compilation, we eliminate the need for separate batch normalization during inference. This optimization consistently reduces computational overhead and improves inference latency across different hardware platforms.}

The second fusion considered is between the convolution and activation functions (ReLU).
The fusion of Conv and ReLU was performed as shown in Eq.~\eqref{eq:foloat_fusion_conv_relu}.
\begin{align}
\text{ReLU}(\text{conv}(x)_i) = \max(0, W_i * X_i + B_i),
\label{eq:foloat_fusion_conv_relu}
\end{align}
where $W_i$, $X_i$, and $B_i$ denote the filter weight, input data, and bias, respectively.
$\max(0, \cdot)$ denotes the ReLU function.
The fusion of convolution and rectified linear units (ReLU) is relatively straightforward, and maintaining scale factors for quantization that enhance accuracy is crucial.
To achieve this, the proposed operator fusion substitutes the output scale of the ReLU with that of the convolution.
In this case, there is no need to maintain the output scale of the convolution, and the quantization process for both the output of the convolution and the input of the ReLU is eliminated, resulting in an improved overall model accuracy and reduced computational overhead.
The substitution of scales and the improvement in accuracy can be expressed mathematically as follows.

\purple{Eq.~\eqref{eq:quant_conv} offers a simplified depiction of the convolution operation by omitting the bias term, focusing on the essential computation. In Eq.~\eqref{eq:quant_conv}, the quantized input feature map $x_i^n$ is multiplied by the quantized weights $w_i^n$, and the products are summed over $n$:
}
\purple{
\begin{align} 
y_{fp32} &= S_x S_w \left( \sum_{n=1}^{N} x_i^n w_i^n \right), \nonumber \ \\ 
y_{i8} &= \text{ROUND} \left( \frac{S_x S_w}{S_y} \sum_{n=1}^{N} x_i^n w_i^n \right), 
\label{eq:quant_conv} 
\end{align}}
\purple{where $S_x$ and $S_w$ denote the scale factors for the input and weights, respectively. The product $S_x S_w$ scales the accumulated sum back to the floating-point domain, resulting in $y_{fp32}$. The quantized output $y_{i8}$ is then obtained by scaling the sum with $\frac{S_x S_w}{S_y}$ and applying the rounding function.}

\purple{
Eq.~\eqref{eq:fusion_conv_relu} illustrates the fusion of the convolution and ReLU operations. 
Initially, the convolution output is passed through the ReLU activation function $F$, scaled by $\frac{S_y}{S_a}$:
}

\purple{
\begin{align} 
a_{i8} &= \frac{S_y}{S_a} F \left( \text{ROUND} \left( \frac{S_x S_w}{S_y} \sum_{n=1}^{N} x_i^{n} w_i^{n} \right) \right) \nonumber \ \\ 
&= \frac{S_y}{S_a} \text{MAX} \left( 0, \text{ROUND} \left( \frac{S_x S_w}{S_y} \sum_{n=1}^{N} x_i^{n} w_i^{n} \right) \right) \nonumber \ \\ 
&= \text{MAX} \left( 0, \text{ROUND} \left( \frac{S_x S_w}{S_a} \sum_{n=1}^{N} x_i^n w_i^n \right) \right) \nonumber \ \\ 
&= \text{MAX} \left( 0, \text{ROUND} \left( \frac{y_{fp32}}{S_a} \right) \right). 
\label{eq:fusion_conv_relu} 
\end{align}
}

\purple{
Upon expanding Eq.~\eqref{eq:fusion_conv_relu}, we note that the scale factor $S_y$ is eliminated, consequently optimizing the calculation. 
In this context, $S_y$ represents the scale of the convolution output before ReLU, and $S_a$ is the scale after the activation function. Due to the properties of ReLU, which outputs non-negative values, the relationship $S_a \geq S_y$ is valid. Fusion enhances accuracy when $S_a < S_y$ by performing quantization only once using $S_a$. This single quantization step more precisely captures the dynamic range of the ReLU output, thereby reducing precision loss.
}

\subsection{Utilizing SQNR delta to Improve Sensitivity Analysis}
The SQNR metric can be affected by cumulative quantization noise as it propagates through the layers. Moreover, relying solely on the SQNR information at individual nodes provides insufficient data for constructing accurate sensitivity lists. Therefore, a metric called SQNR delta is needed to reduce cumulative quantization noise and better understand the interactions between layers in DNN models.. The SQNR delta, as a measure of the relative sensitivity, was computed using  Eq.~\eqref{eq:sqnr_gradient}. 
\purple{Because it considers the difference from the previous layer's SQNR, the SQNR delta can mitigate the impact of cumulative quantization noise. Moreover, since it uses the correlation with adjacent layers rather than the SQNR value of a single layer, a smaller SQNR delta value indicates an increase in quantization noise, while a larger value indicates a decrease in quantization noise. The sensitivity of each layer is determined based on the magnitude of the increase or decrease in SQNR delta.}
\purple{
\begin{equation}
\text{SQNR delta}_{n} = \frac{SQNR_{n} - SQNR_{n-1}}{L_{n} - L_{n-1}}
\label{eq:sqnr_gradient}
\end{equation}
}

\begin{figure*}[!h]
	\centering
	\subfloat[][ResNet18v1]{\label{fig:res18_sqnr_graph}\includegraphics[width=.33\textwidth]{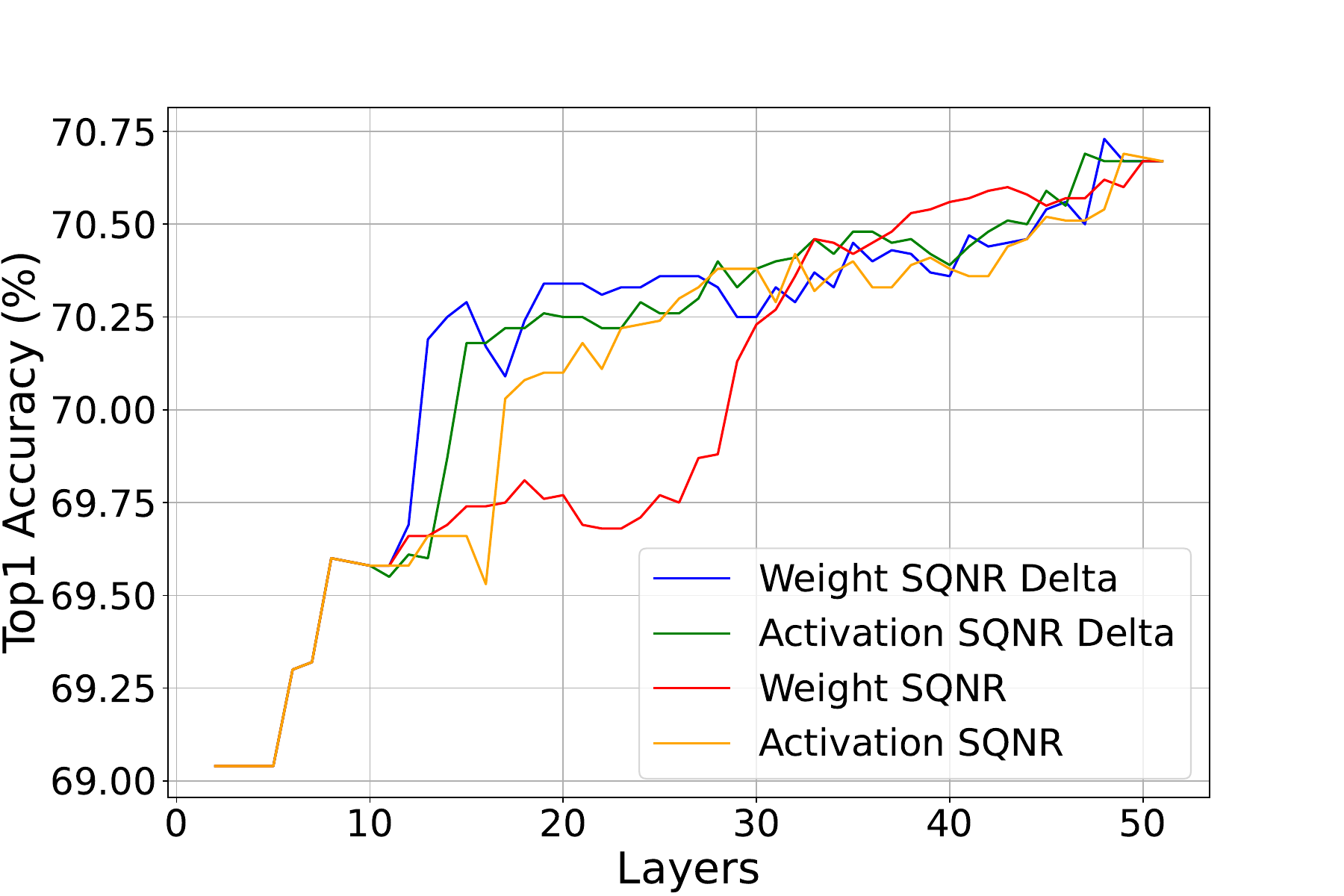}}
	\subfloat[][MobileNetv2]{\label{fig:Mobile_sqnr_graph}\includegraphics[width=.33\textwidth]{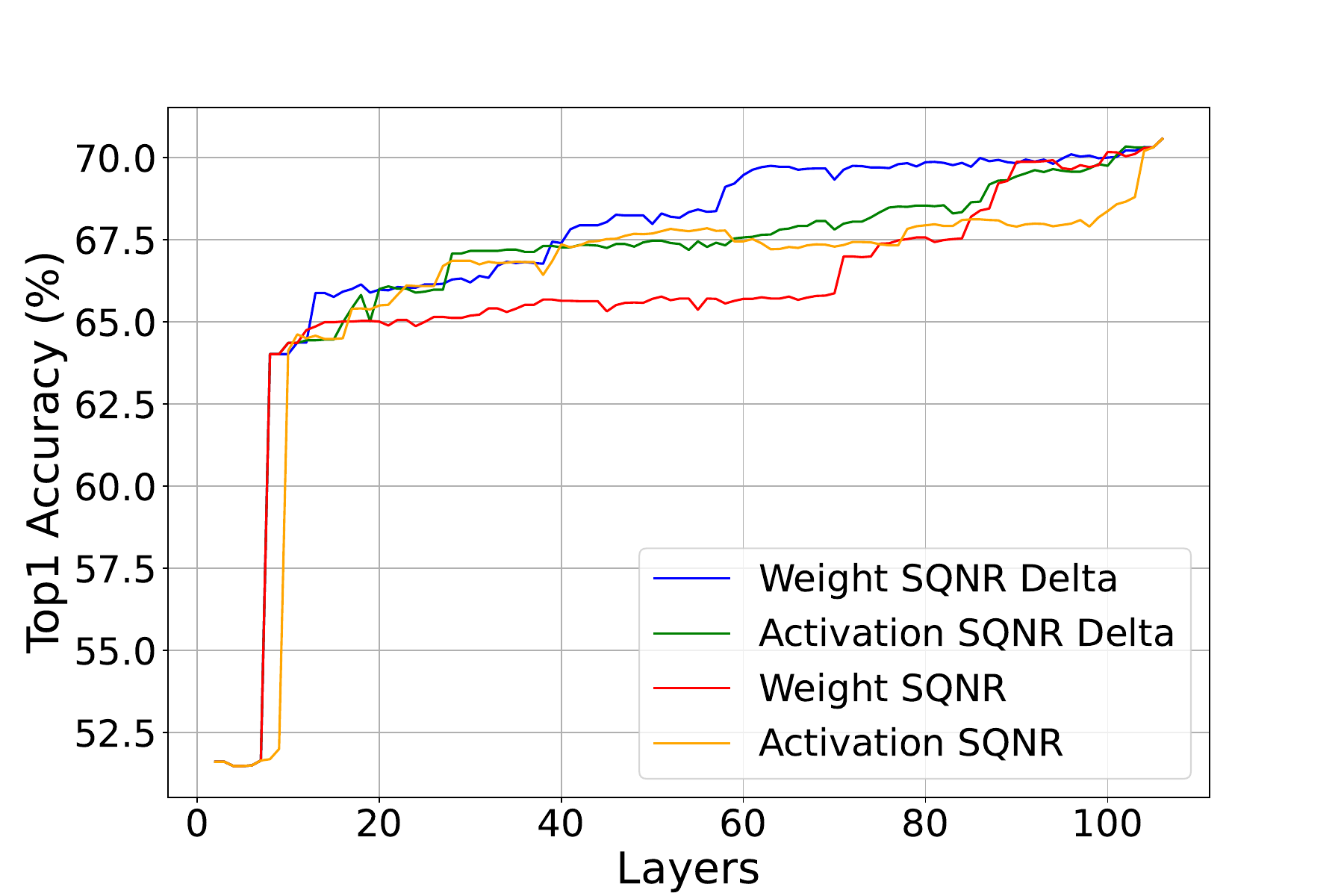}}
	\subfloat[][SqueezeNet]{\label{fig:Squeeze_sqnr_graph}\includegraphics[width=.33\textwidth]{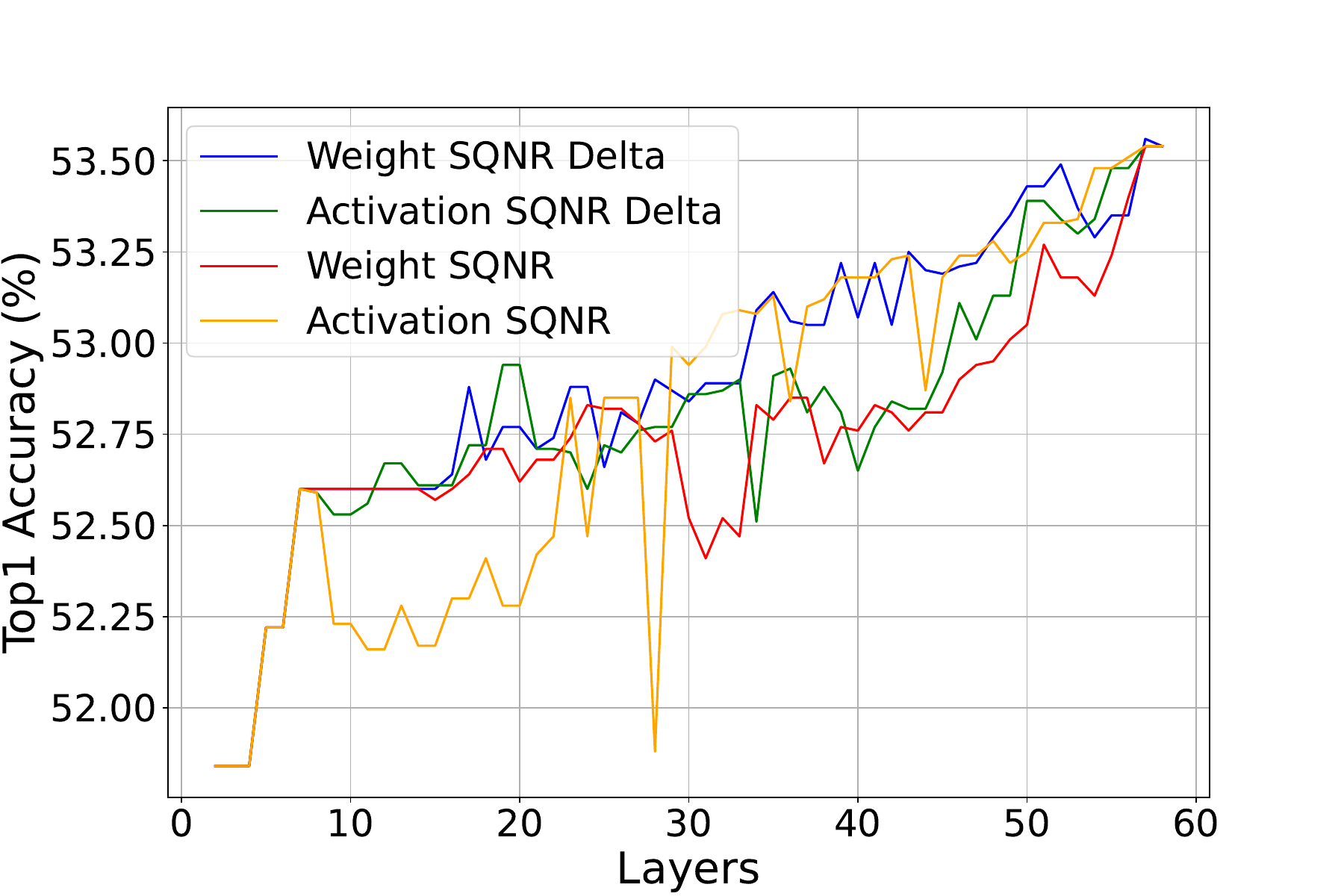}}
    \caption{Comparison of top-1 accuracy changes when dequantizing each layer one by one using SQNR delta and SQNR metrics}
    \label{fig:comp_sqnr}
\end{figure*}
    
Fig.~\ref{fig:comp_sqnr} illustrates the accuracy recovery of the models when each layer is individually dequantized using the sensitivity lists created with weight SQNR delta, activation SQNR delta, weight SQNR, and activation SQNR for ResNet18v1, MobileNetv2, and SqueezeNet.
As shown in Fig.~\ref{fig:comp_sqnr}, the sensitivity lists generated using Weight and Activation SQNR delta recover accuracy approximately 30\% faster than the lists generated solely using SQNR values. In particular cases, the recovery speed was 62\% faster.
Specifically, using the weight SQNR delta for dequantization in ResNet18v1 shows significant accuracy recovery starting from the 13th layer in the sensitivity list. In contrast, using weight SQNR for dequantization shows accuracy recovery starting from the 27th layer. This indicates that SQNR delta provides a better understanding of layer sensitivity than the SQNR metric. While there may be variations depending on the model, activation SQNR delta generally performs dequantization more efficiently than activation SQNR. In particular, in ResNet18v1, activation SQNR delta achieves faster accuracy recovery compared to weight SQNR and demonstrates superior accuracy recovery performance across different DNN models.

\begin{algorithm}[t]
  \caption{Sensitivity list generation for quantization}
  \label{alg:QuantSensitivityList}
  \hspace*{\algorithmicindent}\textbf{Input:} number of layers $n$, quantized layers weight and activations $Q$, dequantized layers weight and activations $D$ \\
  \hspace*{\algorithmicindent}\textbf{Output:} Sensitivity List $S$
  
  \begin{algorithmic}[1]
  \State $SQNRList \gets []$
  \State $MSEList \gets []$
  \State $i \gets 0$
  \While{$i < n$}  \textcolor{blue}{\Comment{Calculating the SQNR, MSE}}
    \State $SQNR \gets$ CalculateSQNR($Q[i]$, $D[i]$)
    \State $MSE \gets$ CalculateMSE($Q[i]$, $D[i]$)
    \State Append $SQNR$ to $SQNRList$
    \State Append $MSE$ to $MSEList$
    \State $i \gets i + 1$
  \EndWhile
  \State $SQNRdeltaList \gets$ CalculateSQNRdeltas($SQNRList$)
  \State $averageMSE \gets$ CalculateAverage($MSEList$)
  \State $S \gets$ RankLayersBySensitivity($SQNRdeltaList$, $MSEList$, $averageMSE$) \textcolor{blue}{\Comment{Generating Sensitivity list}}
  \State \Return $S$
  \end{algorithmic}
\begin{algorithmic}[1]
  \Function{RankLayersBySensitivity}{$G$, $M$, $\bar{M}$}
    \State $L_{\text{mse}} \gets \{m \in M \mid m > 5\bar{M}\}$  
    \State 
    \textcolor{blue}{\Comment{Selecting MSE exceeding 5 times the average MSE}}
    \State $L_{\text{low}} \gets \text{sort}(G, \text{`ascending'})$
    \State
    \textcolor{blue}{\Comment{Ascending sort of SQNR delta values}}
    \For{$l_{\text{mse}}$ in $L_{\text{mse}}$}
      \If{$l_{\text{mse}}$ in $L_{\text{sqnr}}$}
        \State Remove $l_{\text{mse}}$ from $L_{\text{sqnr}}$
        \State Prepend $l_{\text{mse}}$ to $L_{\text{sqnr}}$ 
        \State \textcolor{blue}{\Comment{Prepending $L_{\text{mse}}$ to the top of $L_{\text{sqnr}}$}} 
      \EndIf
    \EndFor
    \State \Return $L_{\text{sqnr}}$ 
  \EndFunction
\end{algorithmic}
\end{algorithm}

\subsection{Sensitivity List Generation}
The final sensitivity list is generated by following the procedure in Algorithm~\ref{alg:QuantSensitivityList}, which incorporates all the considerations mentioned. Algorithm~\ref{alg:QuantSensitivityList} first disables the operator fusion in the DNN model and then calculates the local metrics using a calibration dataset that provides a good understanding of the model.
To ensure a comprehensive sensitivity evaluation for the operators in the DNN model, SQNR was calculated using the weights and the activation tensors.
The computed SQNR is transformed into an indicator called the SQNR delta to reduce the cumulative quantization noise impact and understand the interactions between layers.
Ultimately, combining SQNR delta for weight and activation tensors with MSE, which is advantageous for identifying layers with significant quantization noise, formed a sensitivity list.

This algorithm takes as input the number of layers \( n \) in a DNN model and the pre-quantization (D) and post-quantization (Q) values of the weight and activation tensors for each layer to create a sensitivity list (S) for quantization. It computes the local metrics SQNR and MSE using D and Q for each layer of the DNN model and matches the layers with their corresponding local metrics in a list. Subsequently, using the SQNR values before and after each layer, an SQNR delta list was generated and the average MSE was calculated using the overall MSE value.
The function RankLayersBySensitivity() identifies layers with significantly higher MSE values than the average MSE and records their layer names at the top of the sensitivity list to prioritize them for dequantization. Next, it sorts the lists of weight and activation SQNR delta in ascending order and reorders both lists after a weighted sum. The reason for incorporating additional gain into the weight SQNR delta is experimental evidence, as shown in Fig.~\ref{fig:comp_sqnr}, indicating that the weight SQNR delta better reflects layer sensitivity in models. Therefore, it is given more weight than activation SQNR delta in the sensitivity list. Additionally, if fusion-capable operators are included, their positions can be adjusted to enable operator fusion. Finally, the generated sensitivity list is returned.

\section{Experiments}
To validate the performance of {\systemName}, we conducted the following five experiments:
i) The accuracy recovery effect of mixed-precision quantization;
ii) Comparison of algorithm execution times based on mixed-precision quantization decisions;
iii) Time cost of measuring Top1 accuracy across various hardware;
iv) Comparison of mixed-precision quantization model Inference time using TensorRT;
v) Ablation study of the proposed method.

Each experiment compared {\systemName} with the InOrder, Top1-accuracy based Order, and Weight SQNR Order approaches on five models: ResNet18v1, ResNet50v1, MobileNetv2, VGGNet, and SqueezeNet. 

 \blue{These methods were used as baselines. We selected  prior works, the InOrder and Weight SQNR Order approaches, because they can be applied at compile time, aligning with our focus on methods feasible within compile-time constraints due to the quick execution time required by {\systemName}. Although the Top1-accuracy based Order method cannot be applied at compile time since it requires extensive computations and validation data—it was included to provide a diverse comparison and to evaluate our method against a broader range of approaches. This demonstrates the effectiveness of {\systemName} not only among compile-time feasible methods but also in comparison with methods that may achieve higher accuracy at the expense of longer processing times.}

\blue{\textit{InOrder~\cite{jacob2017}} sequentially applies quantization starting from the first layer of the model as the most basic algorithm.}

\blue{\textit{Weight SQNR Order~\cite{pandey2023practical}} uses only the weight SQNR to generate a sensitivity list ordered by layers with lower SQNR values. This method is applicable at compile times because of its reliance solely on the SQNR metrics, making it practical for identifying layers prone to quantization noise.}

\blue{\textit{Top1-accuracy based Order~\cite{ewha,wu2020integer}} quantizes layers one at a time and evaluates the extent of Top-1 accuracy degradation using validation data. The layer with the largest decrease in accuracy is considered the most sensitive to quantization and forms the Sensitivity List in this order. However, this technique is time-consuming because it directly uses labeled validation data, limiting its applicability during compile time.}

All experiments were conducted on a desktop environment running Ubuntu, with an Intel i7-12700K CPU, 128GB DRAM, and an NVIDIA 3090Ti GPU.
The compiler used was Glow developed by Meta, which utilizes the OpenCL GPU backend to generate the GPU code.




\begin{table*}[ht]
\centering
\caption{\blue{Model accuracy \& size according to the quantization ratio of DNN models (BOPs reduction rate 0\%: original model, 100\%: fully quantized model, Size: model size(MB)). We highlighted the \colorbox[HTML]{CAFFCA}{best value} in green.} }
\label{table:accuracy_summary2}
\resizebox{\textwidth}{!}{
\begin{tabular}{@{}lcccccccccccccc@{}}

\toprule

 \multirow{3}{*}{\ \ \ \ \ \ \ \ \ \ Model} & \multirow{3}{*}{Method} & \multicolumn{10}{c}{BOPs Reduction Rate} \\ \cmidrule(l){3-12} 

 &  & 60\% & Size & 50\% & Size & 40\% & Size &30\%  & Size & 20\%  & Size\\ \midrule

  & \multicolumn{1}{c}{InOrder~\cite{jacob2017}}                   &69.88 &25.81 &69.84 &28.48 &69.97 &33.93 &69.94 &36.70 &70.11 &40.27 \\
  \multicolumn{1}{c}\textbf{ResNet18v1\ } & \multicolumn{1}{c}{Top1-accuracy based Order (50K)~\cite{ewha,wu2020integer}}         &\cellcolor[HTML]{CAFFCA}\textbf{70.21} &25.81 &\cellcolor[HTML]{CAFFCA}\textbf{70.34} &28.49 &\cellcolor[HTML]{CAFFCA}\textbf{70.46} &33.93 &\cellcolor[HTML]{CAFFCA}\textbf{70.49} &35.80 &\cellcolor[HTML]{CAFFCA}\textbf{70.60} &39.47 \\
  \multicolumn{1}{c} \text{(FP32:70.67)} & \multicolumn{1}{c}{\blue{Top1-accuracy based Order (50)}}  &\blue{69.09} &\blue{25.40} &\blue{69.1} &\blue{27.87} &\blue{69.43} &\blue{33.25} &\blue{69.64} &\blue{35.92} &\blue{69.54} &\blue{39.48}\\
  & \multicolumn{1}{c}{Weight SQNR Order~\cite{pandey2023practical}}       &69.33 &25.91 &69.45 &29.47 &69.62 &31.25 &69.95 &36.60 &70.03 &38.38 \\
  & \multicolumn{1}{c}{\textbf{{\systemName} (Ours)}}                      &70.19 &25.81 &70.28 &29.37 &70.31 &31.15 &70.39 &36.50 &70.47 &38.51 \\ \midrule

  & \multicolumn{1}{c}{InOrder~\cite{jacob2017}}                   &74.05 &53.51 &74.28 &61.46 &74.51 &69.96 &74.51 &76.23 &74.46 &83.99 \\
  \multicolumn{1}{c}\textbf{ResNet50v1\ } & \multicolumn{1}{c}{Top1-accuracy based Order (50K)~\cite{ewha,wu2020integer}}         &\cellcolor[HTML]{CAFFCA}\textbf{74.71} &54.22 &74.75 &62.73 &74.9 &69.19 &75.01 &76.39 &75.17 &84.71 \\
  \multicolumn{1}{c} \text{(FP32:75.18)} & \multicolumn{1}{c}{\blue{Top1-accuracy based Order (50)}} &\blue{73.51} &\blue{53.07} &\blue{73.57} &\blue{62.49} &\blue{73.60} &\blue{68.77} &\blue{73.95} &\blue{75.97} &\blue{74.51} &\blue{83.77} \\      
  & \multicolumn{1}{c}{Weight SQNR Order~\cite{pandey2023practical}}               &74.22 &54.59 &74.28 &62.71 &74.29 &68.75 &74.49 &75.45 &74.64 &83.58 \\
  & \multicolumn{1}{c}{\textbf{{\systemName} (Ours)}}                      &74.55 &54.05 &\cellcolor[HTML]{CAFFCA}\textbf{74.81} &61.28 &\cellcolor[HTML]{CAFFCA}\textbf{75.12} &68.52 &\cellcolor[HTML]{CAFFCA}\textbf{75.16} &76.44 &\cellcolor[HTML]{CAFFCA}\textbf{75.19} &82.52 \\ \midrule
              
  & \multicolumn{1}{c}{InOrder~\cite{jacob2017}} &56.76 &10.15 &56.58 &11.70 &57.45 &13.13 &58.01 &14.30 &68.54 &15.82 \\
  \multicolumn{1}{c}\textbf{MobileNetv2} & \multicolumn{1}{c}{Top1-accuracy based Order (50K)~\cite{ewha,wu2020integer}} &\cellcolor[HTML]{CAFFCA}\textbf{68.95} &10.15 &\cellcolor[HTML]{CAFFCA}\textbf{69.2} &11.70 &\cellcolor[HTML]{CAFFCA}\textbf{69.82} &13.13 &\cellcolor[HTML]{CAFFCA}\textbf{70.05} &14.30 &\cellcolor[HTML]{CAFFCA}\textbf{70.29} &15.82\\
  \multicolumn{1}{c} \text{(FP32:70.57)}  & \multicolumn{1}{c}{\blue{Top1-accuracy based Order (50)}} &\blue{65.21} &\blue{10.24} &\blue{65.79} &\blue{11.67} &\blue{65.84} &\blue{13.09} &\blue{66.11} &\blue{14.78} &\blue{67.60} &\blue{16.13}\\       
  & \multicolumn{1}{c}{Weight SQNR Order~\cite{pandey2023practical}} &65.52 &10.05 &65.59 &11.75 &65.70 &13.27 &66.97 &14.39 &68.45 &15.77 \\
  & \multicolumn{1}{c}{\textbf{{\systemName} (Ours)}}  &66.53 &10.18 &66.86 &11.54 &66.88 &13.06 &67.87 &14.52 &68.39 &15.92\\ \midrule

  & \multicolumn{1}{c}{InOrder~\cite{jacob2017}}                   &52.70 &2.60 &53.00 &2.98 &53.24 &3.36 &53.21 &3.45 &53.24 &4.79\\
  \multicolumn{1}{c}\textbf{SqueezeNet\ \ } & \multicolumn{1}{c}{Top1-accuracy based Order (50K)~\cite{ewha,wu2020integer}}         &52.77 &2.67 &52.75 &2.95 &52.75 &3.95 &52.75 &4.03 &53.14 &4.27\\
  \multicolumn{1}{c} \text{(FP32:53.55)}   & \multicolumn{1}{c}{\blue{Top1-accuracy based Order (50)}} &\blue{52.19} &\blue{2.68} &\blue{52.94} &\blue{2.98} &\blue{53.24} &\blue{3.32} &\blue{53.26} &\blue{3.45} &\blue{53.31} &\blue{4.79} \\       
  & \multicolumn{1}{c}{Weight SQNR Order~\cite{pandey2023practical}}               &52.22 &2.59 &52.50 &2.86 &52.50 &2.86 &52.50 &2.86 &52.48 &4.21\\
  & \multicolumn{1}{c}{\textbf{{\systemName} (Ours)}}                      &\cellcolor[HTML]{CAFFCA}\textbf{53.01} &2.71 &\cellcolor[HTML]{CAFFCA}\textbf{53.02} &2.97 &\cellcolor[HTML]{CAFFCA}\textbf{53.35} &3.25&\cellcolor[HTML]{CAFFCA}\textbf{53.43} &3.44 &\cellcolor[HTML]{CAFFCA}\textbf{53.52} &4.79\\ \midrule

  & \multicolumn{1}{c}{InOrder~\cite{jacob2017}}                   &69.44 &255.11 &69.44 &327.88 &69.39 &382.47 &69.48 &418.85 &69.41 &455.23\\
  \multicolumn{1}{c}\textbf{VGGNet \ \ \ }    & \multicolumn{1}{c}{Top1-accuracy based Order (50K)~\cite{ewha,wu2020integer}}         &69.79 &309.45 &69.73 &345.89 &69.80 &392.32 &69.78 &418.70 &69.84 &455.07\\
  \multicolumn{1}{c} \text{(FP32:70.00)}   & \multicolumn{1}{c}{\blue{Top1-accuracy based Order (50)}} &\blue{69.62} &\blue{271.66} &\blue{69.66} &\blue{362.60} &\blue{69.77} &\blue{392.50} &\blue{69.81} &\blue{418.87} &\blue{69.88} &\blue{455.25} \\         
  & \multicolumn{1}{c}{Weight SQNR Order~\cite{pandey2023practical}}               &69.34 &280.42 &69.38 &334.99 &69.37 &371.39 &69.33 &407.78 &69.49 &471.54\\
  & \multicolumn{1}{c}{\textbf{{\systemName} (Ours)}}                      &\cellcolor[HTML]{CAFFCA}\textbf{69.79} &291.30 &\cellcolor[HTML]{CAFFCA}\textbf{69.81} &345.88 &\cellcolor[HTML]{CAFFCA}\textbf{69.86} &391.48 &\cellcolor[HTML]{CAFFCA}\textbf{69.86} &427.85 &\cellcolor[HTML]{CAFFCA}\textbf{69.88} &473.36\\ 
\bottomrule

\end{tabular}
                                                  }
\end{table*}

\begin{figure*}[t]
	\centering
	\subfloat[][\blue{ResNet18v1}]{\label{fig:q_resnet18}\includegraphics[width=.33\textwidth]{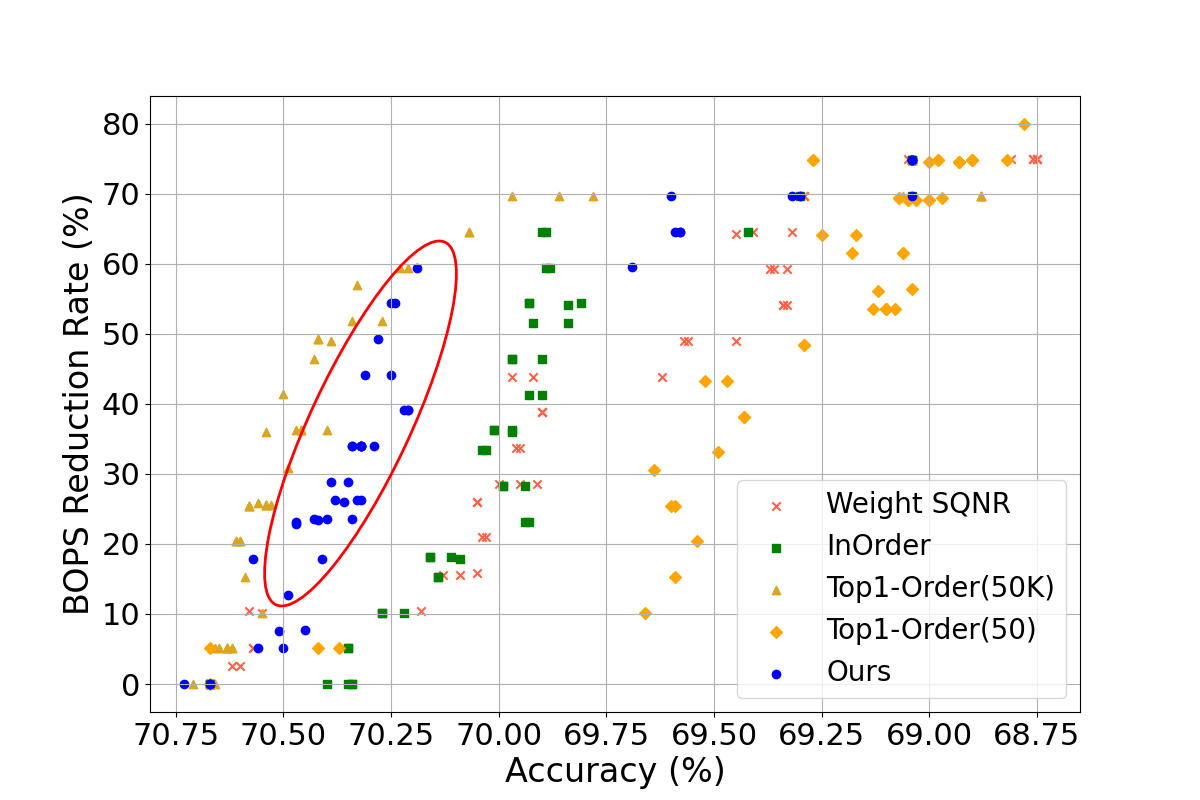}}
	\subfloat[][\blue{ResNet50v1}]{\label{fig:q_resnet50}\includegraphics[width=.33\textwidth]{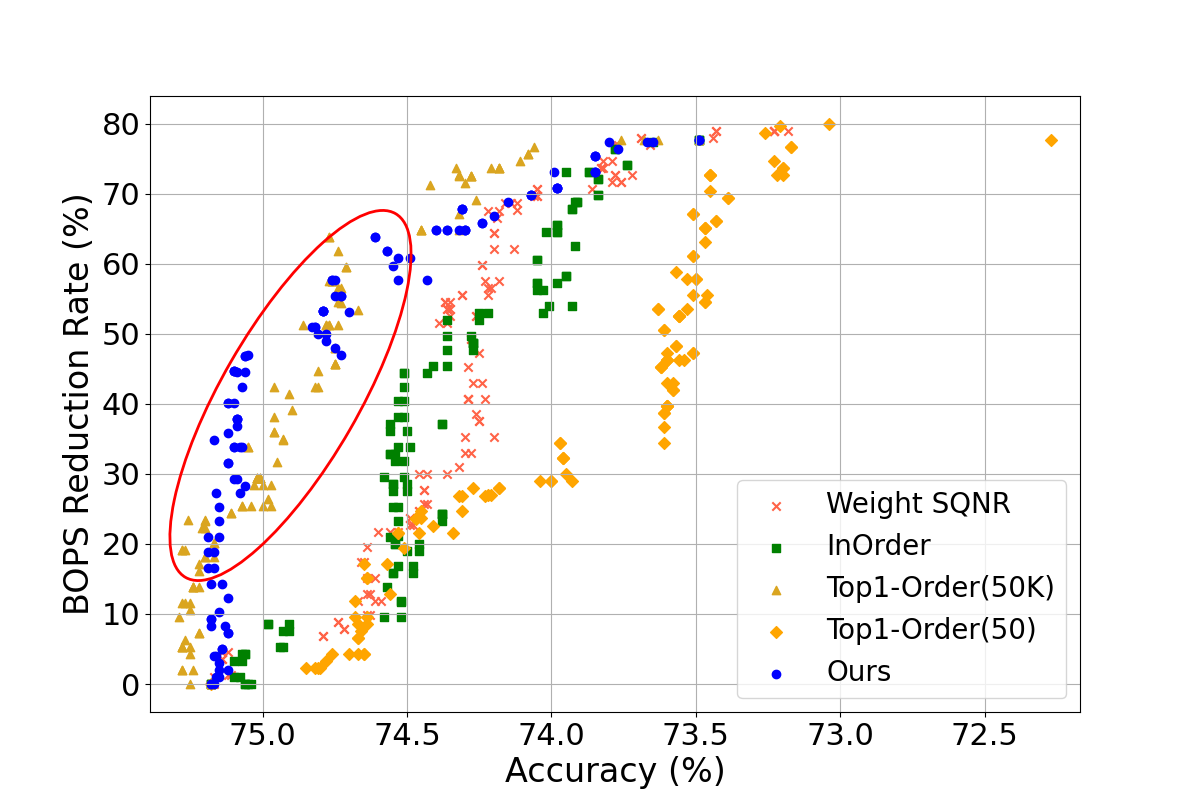}}
	\subfloat[][\blue{MobileNetv2}]{\label{fig:q_mobilenet}\includegraphics[width=.33\textwidth]{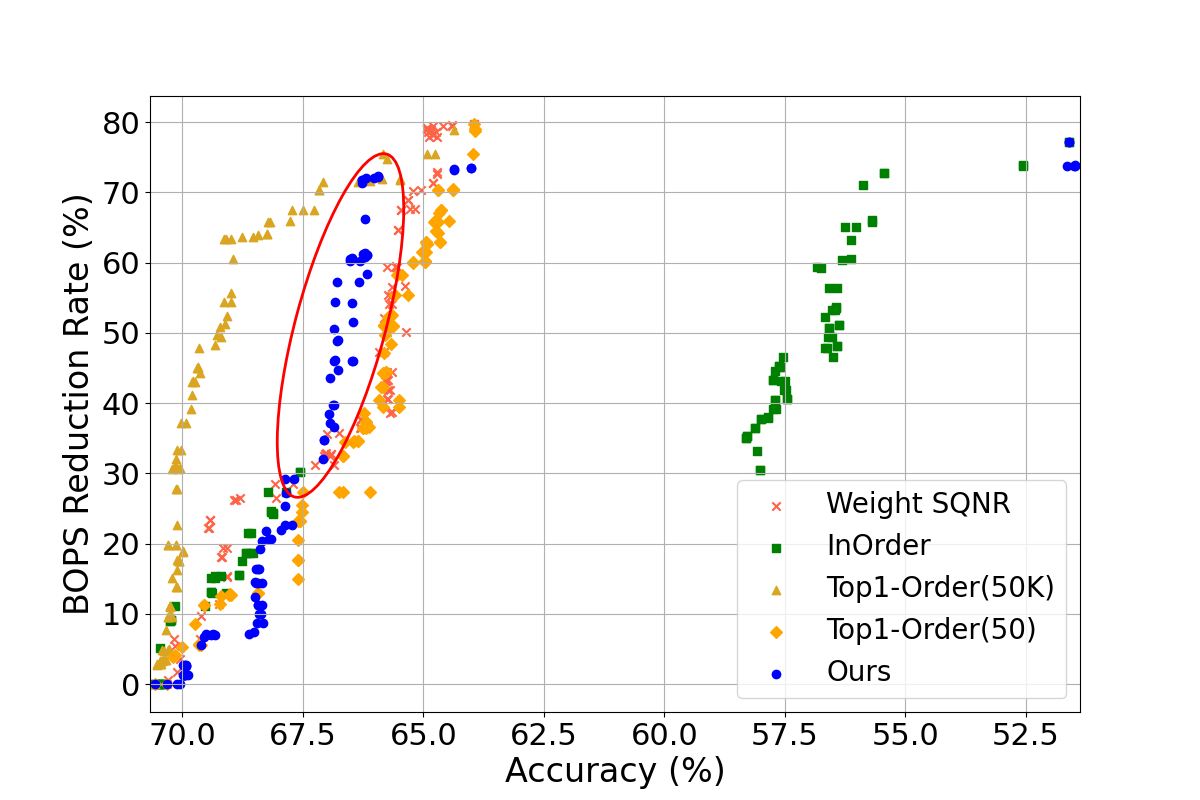}}\vfill
	\subfloat[][\blue{SqueezeNet}]{\label{fig:q_squeezenet}\includegraphics[width=.33\textwidth]{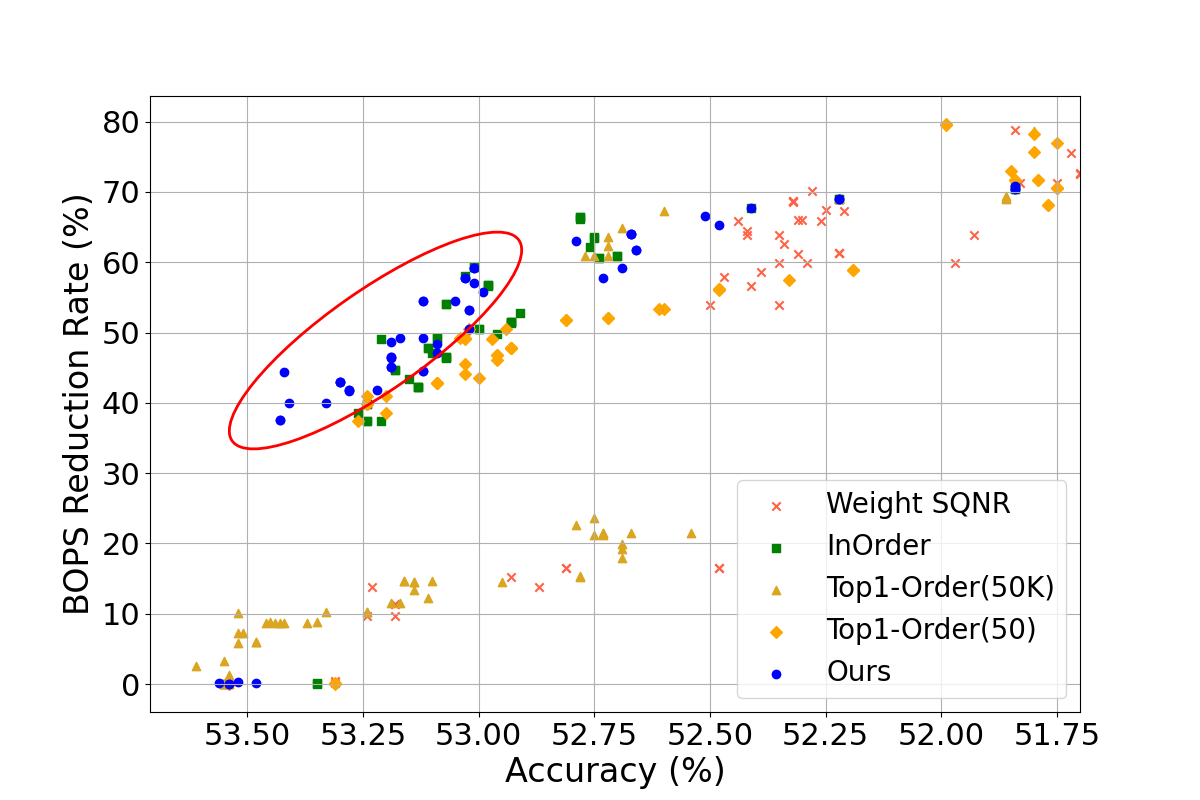}}	
	\subfloat[][\blue{VGGNet}]{\label{fig:q_shufflenet}\includegraphics[width=.33\textwidth]{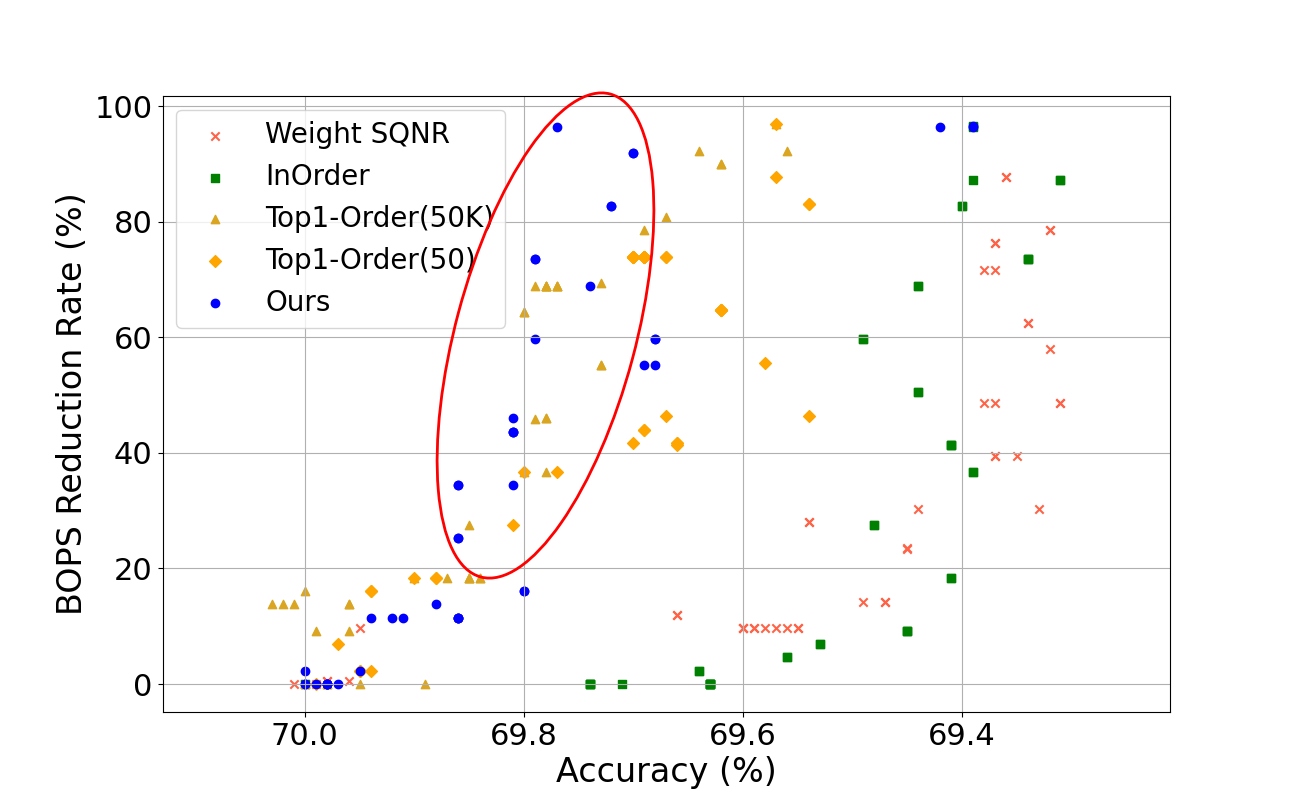}}\vfill

	\caption{\blue{Comparison of model accuracy according to BOPs reduction rate of DNN models (BOPs reduction rate 0\%: original model, 100\%: fully quantized model)}}
	\label{fig:dot_Comparision}
\end{figure*}





\subsection{Comparison Experiment of Mixed-Precision Quantization Methods}
In this experiment, we performed mixed-precision quantization of the DNN models using a Sensitivity List generated based on the local metric. We compared and analyzed the impact on the accuracy and size when quantizing five DNN models using previous methods (three approaches) and our proposed technique (\textit{{\systemName}}), which is based on local metrics and sensitivity list generation.
The experiments were conducted by evaluating mixed-precision quantization according to Eq.~\eqref{eq:bops}, calculating the Bit Operations (BOPs), and determining the BOPs reduction rate due to quantization.

\begin{align}
&\text{BOPs Reduction Rate} = \left( 1 - \frac{\text{BOPs}(\theta)}{\text{BOPs}_{\text{fp32}}} \right) \times 100\%, \nonumber \\
&\text{BOPs}(\theta) = \sum_{ops_i \in \text{network}} bits(\theta_i) \cdot MAC(op_i), \nonumber \\ 
&\theta_i \in {\text{8bit or 32bit}} 
\label{eq:bops}
\end{align}
where $op_i$ denotes operations within the network, bits($\phi_i$) refers to the bit-width related to weight and activations for the operation $op_i$, and $MAC(op_i)$ indicates the total number of Multiply-Accumulate (MAC) operations for operation $op_i$. 
Therefore, $\text{BOPs}(\theta)$ represents the BOPs when applying mixed-precision, and $\text{BOPs}_{\text{fp32}}$ denotes the BOPs of the original model without quantization. 
Finally, the  existing and proposed methods were compared based on accuracy and model size according to the BOPs reduction rate for each model's mixed-precision level.

\blue{Also to provide a fair and practical comparison, we additionally experimented with the Top1-accuracy based Order method using only 50 images, referred to as Top1-accuracy based Order (50).
While the Top1-accuracy based Order (50K) method, which utilizes 50,000 images from the validation set, can achieve slightly higher accuracy in some cases, it is impractical for rapid deployment due to its significant computational and data requirements. Therefore, we simulated a realistic scenario using only 50 images and compared and analyzed the performance of our method and the Top1-accuracy based Order (50) under practical constraints. Moreover, our method does not require any labeled data, whereas the Top1-accuracy based Order methods rely on labeled data even for 50 images. Moreover, our method does not require any labeled data, whereas the Top1-accuracy based Order methods rely on labeled data even when using only 50 images.
As shown in Table~\ref{table:accuracy_summary2}, considering all methods, {\systemName} improved the accuracy of the experimental models compared to the InOrder, Top1-accuracy based Order (50), and Weight SQNR Order methods: 0.83\% for Resnet18v1, 0.73\% for Resnet50v1, 10.28\% for MobileNetv2, 0.55\% for SqueezeNet, and 0.47\% for VGGNet.}


\blue{Fig.~\ref{fig:dot_Comparision} shows accuracy changes for each technique when applying mixed-precision per layer across different models.}

\blue{For the ResNet18v1 model, {\systemName} demonstrated accuracy improvements of up to 1.10\% compared to the Top1-accuracy Order(50) method and up to 0.83\% over the InOrder and Weight SQNR Order methods. Among quantization methods executable within compile time, it consistently showed the best performance.}

\blue{The ResNet50v1 model, with its higher number of layers, exhibited linear accuracy recovery with inverse quantization. {\systemName} showed the best quantization performance with a BOPS reduction rate between 45\% and 75\%,  outperforming the Top-1 accuracy-based Order methods. 
{\systemName} consistently shows better quantization performance across almost all ranges compared with the InOrder, Top1-accuracy based Order (50) and Weight SQNR Order methods.}

\begin{figure}[t]
    \centering
    \includegraphics[width=1\linewidth,scale=0.7]{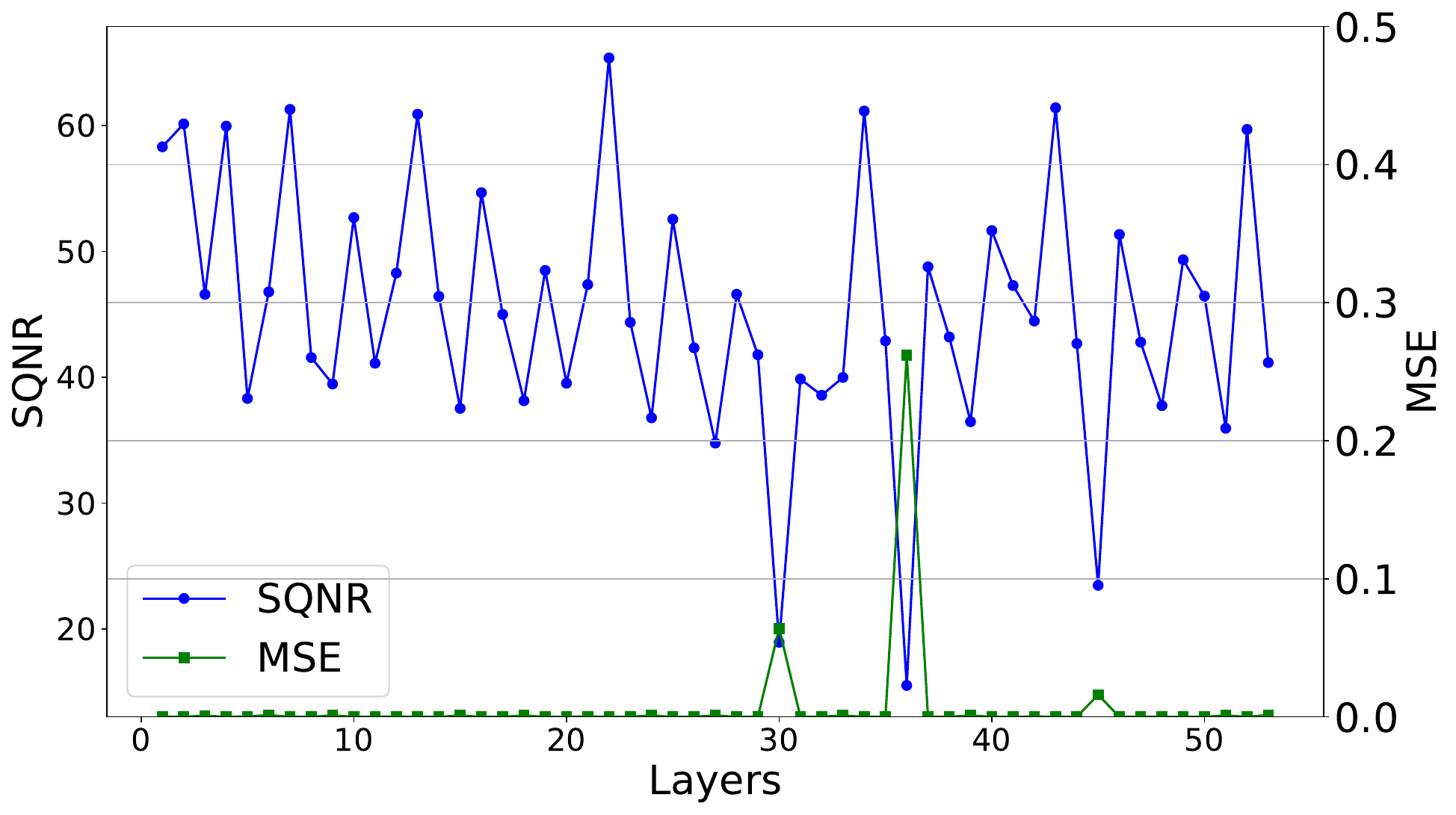}
    \caption{Weight SQNR of MobileNetv2}
    \label{fig:Weight SQNR of MobileNet v1}
\end{figure}

\blue{The MobileNetv2 model, despite having numerous layers, benefits significantly from minor inverse quantization, achieving an accuracy recovery of approximately 12.5\%. {\systemName} outperformed the Top1-accuracy based Order (50) method by up to 2.35\%. Additionally, {\systemName} consistently outperformed the weight SQNR Order and InOrder methods across almost all ranges. MobileNetv2 uses depth-wise separable convolution instead of standard convolution, increasing the number of filters and accumulating quantization noise independently for each channel, leading to significant overall noise accumulation. Fig.~\ref{fig:Weight SQNR of MobileNet v1} illustrates the layer-wise weight SQNR pattern of MobileNetv2, showing a decrease in the SQNR values in depthwise convolution and pointwise convolution, followed by an increase in 1x1 convolution. Owing to the fluctuating quantization noise, identifying layers sensitive to quantization using only weight SQNR Order methods can be challenging. thus, the proposed approach combines activation, weight SQNR delta, and MSE metrics to provide comprehensive layer sensitivity information, achieving up to 1.18\% better accuracy compared to the Weight SQNR Order method.}

\blue{The SqueezeNet and VGGNet models have fewer layers, making an accurate evaluation of the sensitivity of each layer crucial. {\systemName}, which utilizes weight, activation SQNR delta, and MSE metrics, provides a better evaluation of layer sensitivity compared to other methods. It achieves better quantization performance in most ranges compared to the Top1-accuracy based Order (50), Weight SQNR Order, and InOrder methods, with up to 0.31\% higher accuracy recovery performance.
In summary, {\systemName} demonstrates a superior evaluation of DNN model layer sensitivity compared to other methods, demonstrating significant efficiency in  accuracy recovery, compression ratio, and time.}

\begin{table}[H]
\centering
\caption{Comparison of sensitivity list generation times [s] (I: InOrder, T: Top1-accuracy based Order, W: Weight SQNR Order)}
\resizebox{\columnwidth}{!}{
\begin{tabular}{ccccccc}
\toprule
\multirow{2}{*}{Model} & \multicolumn{2}{c}{\multirow{2}{*}{\# of Layers}} & \multicolumn{4}{c}{Method} \\ \cmidrule(l){4-7} 
 &  &  & I & T & W & Ours  \\ \hline

ResNet18v1   &\multicolumn{2}{c}{50}  &  0       & 376,350               & 0.423          & 6.345        \\ 
ResNet50v1   &\multicolumn{2}{c}{123} &  0       & 504,000               & 1.176          & 18.994        \\ 
MobileNetv2  &\multicolumn{2}{c}{105} &  0       & 417,690               & 0.672          & 10.404       \\ 
SqueezeNet &\multicolumn{2}{c}{57}  &  0       & 161,504               & 0.374          & 5.790          \\ 
VGGNet     &\multicolumn{2}{c}{46}  &  0       & 182,300               & 0.210          & 2.990        \\ 
\bottomrule
\end{tabular}
}
\label{table:execution_time}
\end{table}

\subsection{Comparison of Running Times}
We compared the time required to create sensitivity lists using different mixed-precision quantization approaches. Table.~\ref{table:execution_time} shows the time required for each method.
The Inorder (I) method requires no time because it does not create sensitivity lists.
The Top1-accuracy based Order (T) method measures sensitivity by quantifying the degree of accuracy degradation per layer during quantization using a validation set of 50,000 images.
Because validation must be performed for all the layers, creating a sensitivity list requires the longest time.
The Weight SQNR Order (W) method, which uses only the SQNR metric, can generate sensitivity lists the fastest.

In this study, the proposed \textit{{\systemName}} takes longer than the Weight SQNR Order method, but compared to the Top1-accuracy based Order method, both \textit{{\systemName}} and Weight SQNR Order method can be performed within compile time, making the comparison of generation times less significant.
For ResNet18v1, while the Top1-accuracy based Order method takes 376,350 seconds, the proposed method processes it in only 6.345 seconds, showing a 99.99\% reduction in time.
Therefore, the proposed method can achieve efficient mixed-precision quantization in an extremely short time compared to the Top1-accuracy based Order method.
Applying the proposed method to large-scale DNN models will likely maintain relatively more faster mixed-precision quantization and high accuracy performance.

\subsection{\red{Time on Various Hardware for Measuring Top-1 Accuracy}}

\red{To evaluate the performance of {\systemName} on different hardware architectures, we implemented it within the compiler stack to ensure that the quantized models can run on CPUs, GPUs, and ARM devices. This approach allowed us to generate executable binaries targeting a server-side CPU (Intel i7-12700K), an edge device CPU (ARM Cortex-A53), and a server-side GPU (NVIDIA 3090Ti). As shown in Table~\ref{table:measuring_top1}, we measured the Top-1 accuracy of various DNN models with 50\% quantization applied across these hardware platforms.} 


\begin{table}[h]
\centering
\caption{\red{Top-1 Accuracy Measurement Time for DNN Models with 50\% Quantization Applied Across Various Target Devices}}
\resizebox{\columnwidth}{!}{
\begin{tabular}{lccccc}
\toprule
 Model & ResNet18v1 & Resnet50v2 & MobileNetv2 & SqueezeNet & VGGNet \\
\hline
GPU(FP32) &0:56:09 &1:27:35 &1:05:54 &0:51:25 & 1:09:43\\
GPU(Mixed) & 0:53:43 & 1:01:08 & 0:58:43 & 0:49:55 & 1:01:27 \\
x86 CPU & 2:07:41 & 3:09:18 & 1:45:57 & 1:01:24 & 30:38:44 \\
ARM CPU & 37:01:75 & 62:40:02 & 25:06:01 & 22:47:19 & $>$72h \\
\bottomrule
\end{tabular}
}

\label{table:measuring_top1}
\end{table}
\red{The results indicate that the time required to measure Top-1 accuracy varies significantly depending on the hardware platform and model complexity. GPU, known for their efficient handling of intensive computations, exhibited the shortest measurement times for all models.}

\red{CPUs took longer to measure compared to GPUs due to issues with parallel processing and memory bandwidth. Particularly, as the size of the models increased, the time difference became more pronounced due to the absence of large-scale parallel processing capabilities. Nonetheless, QuantuneV2 suggests that it is possible to create mixed-precision quantized models that can operate on CPUs as well.}

\red{Edge devices like the ARM Cortex-A53 had significantly longer measurement times due to their limited computational resources. Measuring ResNet18v1 took approximately 37 hours, and ResNet50v2 exceeded 62 hours. VGGNet required more than 72 hours. This suggests that deploying complex models on resource-constrained devices may be impractical and requires further optimization. Specifically, applying hardware-accelerated APIs during the compilation stage could enable hardware-optimized execution to maximize performance on each device. However, our focus is on generating efficient mixed-precision quantized models, and optimizations that accelerate execution speed by utilizing hardware acceleration instructions on specific platforms are directions for future research.}
  
\red{We also observed differences in measurement times between FP32 models on the GPU and 50\% quantized models. While quantization and dequantization processes introduce some computational overhead, the reduced bit-width of quantized models offers advantages in execution speed. Based on this, we confirmed that the mixed-precision quantization model using {\systemName} performed 2.92\% to 30.2\% faster than the FP32 model.}
\red{In conclusion, our experiments demonstrate that {\systemName} operates efficiently across various hardware architectures and has practical utility in embedded AI applications.}

\subsection{\red{Inference Latency using TensorRT}}
\red{To measure the inference speed of the quantized models in {\systemName}, we used TensorRT ~\cite{tensorrt}. TensorRT optimizes 8-bit operations by leveraging NVIDIA's DP4A instructions, which accelerate 8-bit multiply-accumulate operations to optimize inference speed on GPUs. First, using {\systemName}, we generated a sensitivity list of the model layers, selected quantization ratios, specified the layers to quantize using TensorRT's setPrecision() API, and then applied quantization.}

\red{Table.~\ref{tab:Inference_time} shows the average inference times after applying mixed-precision quantization in TensorRT based on the sensitivity analysis results from {\systemName}. When the BOPs reduction rate is 0\%, it represents the original model (32-bit). As the BOPs reduction rate increases, the inference speed improves. This is because the increased utilization of NVIDIA DP4A's 8-bit instructions achieves optimized computation speed. For example, in the case of ResNet18v1, for every 10\% increase in the quantization ratio, there was an average improvement of approximately 9.95\% in inference speed.}
\begin{table}[H]
\centering
\caption{\red{Comparison of speed improvements for the DNN model based on BOPs reduction rates using QuantuneV2 with 10 images}}
\label{tab:Inference_time}
\resizebox{\columnwidth}{!}{
\begin{tabular}{ccccccc}
\toprule
 
  \multirow{2}{*}{Model}  & \multicolumn{6}{c}{BOPs Reduction Rate} \\ \cmidrule(l){2-7} 
     &\textbf{100\%} & \textbf{60\%} & \textbf{50\%}   & \textbf{40\%} & \textbf{20\%} & \textbf{0\%} \\ \hline

\multirow{1}{*}{ResNet18v1}  &2.87x &1.77x  &1.44x    &1.32x  &1.1x  & 1,012\si{\micro\second} \\  
\multirow{1}{*}{ResNet50v1}  &2.75x &1.36x  &1.24x    &1.17x  &1.10x  & 1,788\si{\micro\second} \\  
\multirow{1}{*}{MobileNetv2} &1.56x &1.18x  &1.12x    &1.05x  &1.04x  & 597\si{\micro\second} \\ 
\multirow{1}{*}{SqueezeNet}  &1.84x &1.60x  &1.46x    &1.32x  &1.11x  & 451\si{\micro\second} \\  
\multirow{1}{*}{VGGNet}      &4.10x  &1.75x  &1.56x    &1.27x  &1.05x  & 3,326\si{\micro\second}\\ 
\bottomrule
\end{tabular}
}
\end{table}

\red{Comparing different models, lightweight models like ResNet18 and MobileNet showed relatively smaller increases in inference time as quantization progressed. On the other hand, for complex models such as VGGNet and ResNet50v1, the increase in inference time was more significant at lower BOPs reduction rates, with VGGNet experiencing a sharp increase in inference time from 810μs to 3326μs when transitioning from a 100\% to a 0\% BOPs reduction rate. This indicates that the performance improvement due to quantization is more pronounced in larger DNN models.}
\red{These results demonstrate that combining {\systemName}’s sensitivity analysis with hardware-specific instructions can maximize the performance of quantized models, showing that higher BOPs reduction rates lead to successful model compression and accelerated inference speed. {\systemName} effectively utilizes hardware resources through this process, proving its ability to significantly enhance performance by applying the optimal quantization stages.}
\begin{table}[H]
\centering
\caption{\news{Computation Overhead Time (\si{\micro\second}) per Operator through Quantization-Dequantization in ResNet50v1 (C : Convolution, B : Batch Normalization, A : Add, R : Relu, All : C+B+A+R)}}
\label{tab:QDQ_time}
\resizebox{0.8\columnwidth}{!}{ 
\begin{tabular}{cccccc}
\toprule
 Tensor Shape &\textbf{C+B} & \textbf{A} & \textbf{R} & \textbf{All} & \textbf{Fusion} \\
\hline
\multirow{1}{*}{[1, 256, 56, 56]}  &341 &18 &5 &364 &341\\
\multirow{1}{*}{[1, 512, 28, 28]}  &368 &6  &3  &377 &357\\
\multirow{1}{*}{[1, 1024, 14, 14]} &536 &3 &3 &542 &518\\
\multirow{1}{*}{[1, 2048, 7, 7]}   &1518 &3 &3  &1524 &1458\\
\bottomrule
\end{tabular}
}
\end{table}

\news{Additionally, Table.~\ref{tab:QDQ_time} shows the time consumption for quantization and dequantization of individual operators. Quantizing and dequantizing each operator individually takes considerable time. However, performing quantization with operator fusion reduces the average computation time by approximately 6.32\% compared to quantizing and dequantizing each operator separately.To minimize computational overhead, {\systemName} is designed to combine adjacent convolution, batch normalization, Add, and ReLU operators as much as possible when generating the sensitivity list, enabling them to be set to the same bit-width.}

\begin{table}[H]
\centering
\caption{\red{Average inference time (μs) after applying a 50\% BOPs reduction rate to ResNet50v1 (T : Top-1 Accuracy based Order)}}
\label{tab:OT_time_Comp}
\resizebox{0.5\columnwidth}{!}{ 
\begin{tabular}{ccc}
\toprule
 Model &\textbf{Ours} & \textbf{T} \\
\hline
\multirow{1}{*}{ResNet18v1}  &698 &715  \\
\multirow{1}{*}{ResNet50v1}  &1,435  &1,493  \\
\multirow{1}{*}{MobileNetv2} &532 &548  \\
\multirow{1}{*}{SqueezeNet}  &308 &320  \\
\multirow{1}{*}{VGGNet}      &2,125  &2,429  \\
\bottomrule
\end{tabular}
}
\end{table}

\red{As a result, as shown in Table.~\ref{tab:OT_time_Comp}, when comparing the average inference time after performing 50\% quantization on ResNet50v1 using {\systemName} and the Top-1 Accuracy based Order method, our approach was found to operate 5.09\% faster on average, with a maximum improvement of 12.52\%. In particular, for larger models, {\systemName} demonstrated faster inference speeds. This was attributed to fewer quantization-dequantization operations in {\systemName}, with an average of four fewer such operations compared to the Top-1 Accuracy based Order method.}

\subsection{Ablation Study}
We conducted an ablation study to evaluate the impact of the local metric indicators selected for mixed-precision quantization on the performance of the DNN models, as listed in Table~\ref{table:ablation_test}.
The comparative experiments were conducted independently across each DNN model, performing mixed-precision quantization in the order of Weight SQNR, IR Opt., SQNR delta, Weight and Activation SQNR delta Mixup, and MSE Mixup. The mixed-precision quantization application rate was set at 40\%.

\begin{table}[H]
\centering
\caption{Ablation studies on mixed-precision quantization method for DNN models(quantization at 40\%), where \cmark indicates that the component is considered. The \colorbox[HTML]{CAFFCA}{best value} is highlighted in green.}
\label{table:ablation}
\resizebox{\columnwidth}{!}{
\begin{tabular}{>{\centering\arraybackslash}m{1.5cm}ccccccc}
\toprule

\multicolumn{1}{c}{\textbf{Model}}  & \multicolumn{1}{c}{\textbf{W SQNR Order}} & \multicolumn{1}{c}{\textbf{IR Opt.}} & \multicolumn{1}{c}{\textbf{W SQNR delta}} & \multicolumn{1}{c}{\textbf{W\&A SQNR Mixup}} & \multicolumn{1}{c}{\textbf{MSE Mixup}} & \multicolumn{1}{c}{\textbf{Accuracy}} 
\\   

\midrule
\multirow{5}{*}{ResNet18v1} 
 & \cmark & \xmark & \xmark & \xmark & \xmark  &  69.62\%\\ 
 & \xmark & \cmark & \xmark & \xmark & \xmark  &  69.84\%\\ 
 & \xmark & \xmark & \cmark & \xmark & \xmark  &  69.82\%\\
 & \xmark & \xmark & \xmark & \cmark & \xmark  &  70.18\%\\
 & \xmark & \xmark & \xmark & \xmark & \cmark  &  \colorbox[HTML]{CAFFCA}{70.31\%}\\

\midrule
\multirow{5}{*}{ResNet50v1}
  & \cmark & \xmark & \xmark & \xmark & \xmark &  74.29\% \\ 
  & \xmark & \cmark & \xmark & \xmark & \xmark &  74.61\% \\  
  & \xmark & \xmark & \cmark & \xmark & \xmark &  74.89\% \\
  & \xmark & \xmark & \xmark & \cmark & \xmark &  75.01\%  \\
  & \xmark & \xmark & \xmark & \xmark & \cmark &  \colorbox[HTML]{CAFFCA}{75.12\%} \\

\midrule
\multirow{5}{*}{MobileNetv2}  
  & \cmark & \xmark & \xmark & \xmark & \xmark & 65.70\% \\  
  & \xmark & \cmark & \xmark & \xmark & \xmark & 66.44\%  \\ 
  & \xmark & \xmark & \cmark & \xmark & \xmark & 66.47\%  \\
  & \xmark & \xmark & \xmark & \cmark & \xmark & 66.66\% \\
  & \xmark & \xmark & \xmark & \xmark & \cmark & \colorbox[HTML]{CAFFCA}{66.88\%} \\

\midrule
\multirow{5}{*}{SqueezeNet} 
  & \cmark & \xmark & \xmark & \xmark & \xmark &  52.48\% \\
  & \xmark & \cmark & \xmark & \xmark & \xmark &  52.66\% \\ %
  & \xmark & \xmark & \cmark & \xmark & \xmark &  53.01\% \\
  & \xmark & \xmark & \xmark & \cmark & \xmark &  53.28\% \\
  & \xmark & \xmark & \xmark & \xmark & \cmark &  \colorbox[HTML]{CAFFCA}{53.35\%} \\

\midrule
\multirow{5}{*}{VGGNet}  
  & \cmark & \xmark & \xmark & \xmark & \xmark & 69.37\% \\ 
  & \xmark & \cmark & \xmark & \xmark & \xmark & 69.74\%\\ %
  & \xmark & \xmark & \cmark & \xmark & \xmark & 69.76\%  \\
  & \xmark & \xmark & \xmark & \cmark & \xmark & 69.79\%  \\
  & \xmark & \xmark & \xmark & \xmark & \cmark & \colorbox[HTML]{CAFFCA}{69.86\%} \\

\bottomrule
\end{tabular}
}
\label{table:ablation_test}
\end{table}

Table~\ref{table:ablation_test} highlights four key findings:
First, using Weight SQNR delta was more effective than using Weight SQNR across five DNN models, resulting in performance improvements ranging from 0.2\% to as much as 0.77\%.

Second, IR Opt. represents the results of calculating the local metrics after the operators have been fused and applying mixed-precision quantization as proposed in this study. 
While it showed an average accuracy improvement of 0.37\% compared to using Weight SQNR, the model's accuracy was not consistently maintained at a high level because the sensitivity to operators could not be accurately assessed.

Third, starting from weight SQNR delta, we unfused the operators to compute local metrics. Weight SQNR delta was found to be more effective than Weight SQNR alone and combining the activation SQNR delta with weight SQNR delta resulted in an approximate 0.2\% accuracy improvement.
Lastly, incorporating the MSE Mixup, as proposed in this study, resulted in the highest model accuracy. In particular, MobileNetv2 showed the largest accuracy improvement of approximately 1.18\% with the addition of MSE Mixup.

Through this ablation study, we confirmed that using the weight and activation SQNR delta along with the MSE mixup is most effective in improving the performance across most models. Significant performance improvements were observed, particularly in networks with more layers, such as ResNet50v1 and MobileNetv2, and meaningful improvements were observed in shallower models, such as ResNet18v1 and VGGNet.
Applying these components to each DNN model (ResNet18v1, ResNet50v1, MobileNetv2, SqueezeNet, and VGGNet) through an ablation study validated that the proposed approach outperformed existing methods.

\section{Conclusion}
\blue{In this study, we proposed {\systemName}, a compiler-based mixed-precision quantization solution aimed at minimizing accuracy degradation due to quantization. To construct sensitivity lists, we selected SQNR delta and MSE as local metrics and evaluated mixed-precision quantization on five DNN models, demonstrating precise identification of layer sensitivities. This accuracy allowed for an efficient balance between model size and accuracy. Additionally, by fusing convolution, batch normalization, and ReLU operators, we reduced computational overhead. This operator fusion achieved faster inference times, particularly for heavily quantized models, with a maximum improvement of approximately 12.53\%.}

\blue{We verified {\systemName} across various hardware platforms, such as GPU, CPU and ARM, confirming its efficiency and applicability in different deployment scenarios, including embedded AI applications. {\systemName} supports mixed-precision quantization during compile time without the need for retraining, enabling rapid deployment and reduced complexity in resource-constrained environments.}

\section*{Acknowledgments}
This work was supported by the Institute of Information \& Communications Technology Planning \& Evaluation(IITP) grant funded by the Korea government(MSIT) (No.RS-2024-00459797, Development of ML compiler framework for on-device AI) and (No.RS-2023-00277060, Development of open edge AI SoC hardware and software platform).

\bibliography{elsarticle-template}


\end{document}